\documentclass[conference]{IEEEtran}
\usepackage{graphicx}
\usepackage{amsmath}
\usepackage{booktabs}
\usepackage[ruled,vlined]{algorithm2e}
\usepackage{subcaption}

\IEEEoverridecommandlockouts

\renewcommand{\footnoterule}{%
  \kern -3pt
  \hrule width 0.12\textwidth
  \kern 2.6pt
}

\begin{document}

\title{Why Do Neural Networks Forget: A Study of Collapse in Continual Learning}

\author{
\IEEEauthorblockN{Yunqin Zhu}
\IEEEauthorblockA{
Department of Electrical and Computer Engineering\\
University of Alberta\\
Edmonton, Canada\\
\texttt{yunqin@ualberta.ca}
}
\and
\IEEEauthorblockN{Jun Jin}
\IEEEauthorblockA{
Department of Electrical and Computer Engineering\\
University of Alberta\\
Edmonton, Canada\\
\texttt{jjin5@ualberta.ca}
}
\thanks{This work is based on a capstone project report submitted in partial fulfillment of the requirements for the MEng degree in the Department of Electrical and Computer Engineering at the University of Alberta.}
}

\maketitle

\begin{abstract}
\noindent Catastrophic forgetting is a major problem in continual learning, and lots of approaches arise to reduce it. However, most of them are evaluated through task accuracy, which ignores the internal model structure. Recent research suggests that structural collapse leads to loss of plasticity, as evidenced by changes in effective rank (eRank). This indicates a link to forgetting, since the networks lose the ability to expand their feature space to learn new tasks, which forces the network to overwrite existing representations. Therefore, in this study, we investigate the correlation between forgetting and collapse through the measurement of both weight and activation eRank. To be more specific, we evaluated four architectures, including MLP, ConvGRU, ResNet-18, and Bi-ConvGRU, in the split MNIST and Split CIFAR-100 benchmarks. Those models are trained through the SGD, Learning-without-Forgetting (LwF), and Experience Replay (ER) strategies separately. The results demonstrate that forgetting and collapse are strongly related, and different continual learning strategies help models preserve both capacity and performance in different efficiency.
\end{abstract}


\section{Introduction}
\noindent Continual learning, also known as incremental learning or lifelong learning, is a study that focuses on training models on a sequence of tasks while preserving the knowledge of previously learned tasks. Since continual learning is characterized by learning from dynamic data distributions, most models suffer from the catastrophic forgetting problem~\cite{wang2024comprehensivesurveycontinuallearning}, where performance on old tasks degrades after training on new tasks. This results from overwriting the internal representations for old tasks while the updates are needed to learn new tasks. This overwriting compresses the internal features, which causes representational collapse and loss of model plasticity. Because the ability to grow and adapt over time is critical for real-world applications, forgetting problem is a serious obstacle. In this study, we demonstrate that catastrophic forgetting is a geometric failure caused by structural and representational collapse, and that the effective rank (eRank) can be used to monitor this process.

To solve catastrophic forgetting, many strategies are developed, such as the replay experience buffer, which stores and reuses previous knowledge; regularization-based methods, which penalize updates on the important weights for previous tasks; and distillation-based methods, which preserve the behavior of old tasks during learning~\cite{Kirkpatrick_2017}~\cite{mallya2018packnetaddingmultipletasks}~\cite{rolnick2019experiencereplaycontinuallearning}. Although these methods reduce the problem of forgetting, the mechanisms behind these successes are still poorly understood. Most prior work evaluated whether these strategies are successful through task performance metrics, especially task accuracy, which limits our understanding of how these methods affect internal representations and model plasticity.

This lack of understanding motivates us to investigate further into representations during training. Recent work indicated that catastrophic forgetting may not be caused by classifier drift or conflicting gradients, but rather by a structural phenomenon called representational collapse. Representational collapse occurs when the model’s internal feature space learned from the network shrinks dramatically to a low-dimensional subspace. In this situation, the network loses its ability to separate and encode new information directions, so the representation is too fragile to incorporate new tasks without interfering with old ones~\cite{nature2024}. 

This may implies the correlation between forgetting and the collapse of the representation space, so the eRank is introduced, a metric used to quantitatively measure the richness and diversity of the representation space. The effective rank quantifies the complexity of the network’s weight matrices and activation representations. High eRank suggests that the information is spread in many directions, so the representation is rich and not redundant. In contrast, low eRank indicates that the information is compressed in a few directions, so the representation is simpler and more redundant~\cite{effrank}.

During our work, we observed that eRank can quantify the complexity of both weight matrices and activation representations, which can measure both structural collapse in the weights and representational collapse in the activations. These two perspectives provide a more complete view of how forgetting is related to the collapse of the model’s internal subspace in continual learning. Therefore, we investigate how eRank, measured for both weights and activations, changes between tasks, architectures, and learning strategies, and how these changes relate to catastrophic forgetting. To better understand the mechanisms, our work also focuses on the following three specific directions.

Across the different architectures, the main focus is on how eRank changes and how it relates to catastrophic forgetting. If forgetting occurs when the representation shrinks to a lower dimension, then the networks experience stronger collapse and severe forgetting. The performance of different architectures provides evidence that collapse is the cause of forgetting. Experience replay (ER) and Learning without Forgetting (LwF) are two common methods to reduce forgetting. Therefore, analyzing their effect on eRank can further investigate the correlation and determine whether the methods can maintain representational diversity. Furthermore, recurrent networks are different from other feedforward architectures in continual learning because they incorporate temporal structure and implicit memory. Investigating the eRank-forgetting correlations in these networks gives a better understanding of the collapse of convolutional and residual networks.
\section{Background}
\begin{figure*}[t]
\centering
\includegraphics[width=\textwidth]{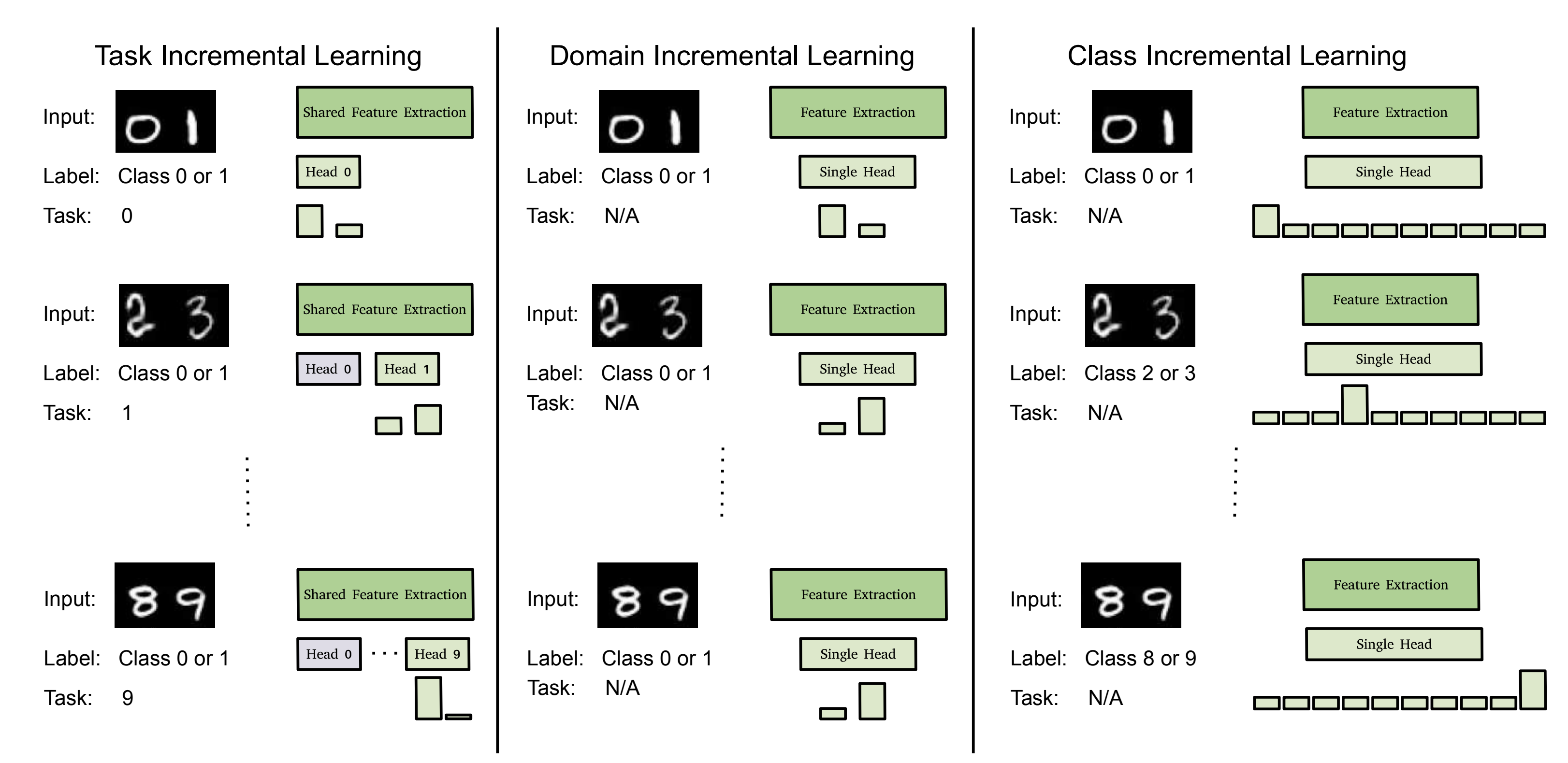}
\caption{Three Different CL Settings~\cite{setting}}\label{fig:clsetting}
\end{figure*}
\noindent Continual learning (CL) is a machine learning paradigm where the models are trained on a sequence of tasks or data over time~\cite{10.3389/fnbot.2021.658280}. Unlike standard supervised learning, where the training data are assumed to be independent and identically distributed, the models are built based on knowledge of past tasks while gaining knowledge of new tasks. This section reviews the core concepts underlying our work, including CL settings, the mechanisms of catastrophic forgetting, and strategies to mitigate it. Then, the architectures, including MLPs, ResNet-18, and two recurrent models, will be outlined.
\subsection{Continual Learning Settings}
\noindent CL is often categorized into three settings, which are task-incremental learning (Task-IL), class-incremental learning (Class-IL), and domain-incremental learning. These settings are different in how tasks are defined, how labels are shared, and whether task identity is available, as shown in Figure~\ref{fig:clsetting}. In this work, two CL settings (Task-IL and Class-IL) are relevant to the experimental design.
\subsubsection{Task-Incremental Learning (Task-IL)}
In Task-IL, the model learns tasks in a sequential way. Each task contains a disjoint subset of classes and has its own output head. The task identity is available only during testing, so the model does not need to distinguish which task a sample belongs to. Since the output layer is protected by the task identities, the drops in performance must be happening deeper in the networks, which can focus the study of forgetting on representation drift. The main challenge in this setting is to preserve meaningful features from old tasks while gaining new ones.
\subsubsection{Class-Incremental Learning (Class-IL)}
In contrast, the model learns new classes over time using a single shared output head in Class-IL. The task identity is not available during testing, and the model must include all learned classes in a single feature space~\cite{ilthreetypes}. This setting is more prone to forgetting because later-learned classes need to be integrated into a global decision space, thereby introducing inter-class competition for representational capacity. The decline in dimensional richness leads to overlap of different class features, which makes the model sensitive to representational collapse.

\subsection{Catastrophic Forgetting}
\noindent Catastrophic forgetting occurs when a model loses performance on old tasks as it learns new ones, because gradient-based updates during learning overwrite the representations learned from older tasks. Forgetting becomes worse when learning more tasks, especially when those tasks are very different. In major studies, forgetting is caused by conflicting gradients, which state that gradients for new tasks increase the loss on old tasks. In other words, the new gradients conflict with the safe region for the old tasks~\cite{Kirkpatrick_2017}. Another major reason is classifier drift, where forgetting occurs in the final layers of the network~\cite{liu2023teamworkgoodempiricalstudy}. When the model learns a new task, it updates the output layer in order to classify the new classes. However, these updates pull the existing decision boundaries, the hyperplane in feature space that separates different classes, towards the new classes. This results in the boundaries that separate older classes becoming distorted.

\subsubsection{Loss of Plasticity}
While conflicting gradients and classifier drift are immediate causes of forgetting, the loss of plasticity is the long-term issue arising from continual learning itself. Plasticity refers to a model's ability to create new representations and expand its feature space. Loss of plasticity is a structural change within the network. As learning more tasks, internal representations become more constrained and collapse into a low-dimensional subspace~\cite{nature2024}. It makes the model lose the ability to adapt its features and create new dimensions for new knowledge. Eventually, the model forgets old knowledge and slowly learns new knowledge.
\subsection{Parameter and Functional Regularization}
\noindent In machine learning, regularization is used to prevent overfitting. However, its function is to prevent catastrophic forgetting in CL as the model is sequentially trained on tasks. When the model learns a new task, the weights are changed to reduce the loss for that task. Regularization adds a penalty that pushes the weights back toward the values that worked well on past tasks~\cite{li2017learningforgetting}, helping to keep the knowledge learned from old tasks stable.  These methods do not store the old knowledge. Regularization methods typically have two groups:
\subsubsection{Parameter-based Regularization}
These methods estimate how important each parameter is for past tasks and then penalize the updates on the parameters that are most critical. As some parameters have task-specific knowledge, this way can keep those parameters close to their old values to reduce forgetting. Two methods, elastic weight consolidation (EWC) and memory aware synapses (MAS), are common in applications. EWC uses the Fisher Information Matrix to find these important parameters, while MAS measures importance through checking the sensitivity of the model’s output~\cite{Kirkpatrick_2017}. These methods perform well in small models, but estimating the importance of the parameters becomes difficult and unreliable in large models or long task sequences.
\subsubsection{Functional Regularization}
These methods focus on the output of the model rather than its parameters. Compared with preventing weight changes, these ensure that the model keeps making similar predictions for old tasks while learning new ones. To achieve this, at least two models are used during training. The teacher model has only old knowledge, and the current model (the student) is encouraged to match the teacher's predictions~\cite{li2017learningforgetting}. The benefit of this is to stabilize representations since matching the old outputs needs the model to retrieve earlier feature directions.
\subsubsection{Learning without Forgetting (LwF)}
\begin{figure*}[t]
\centering

\begin{subfigure}{0.45\textwidth}
  \centering
  \includegraphics[width=\linewidth]{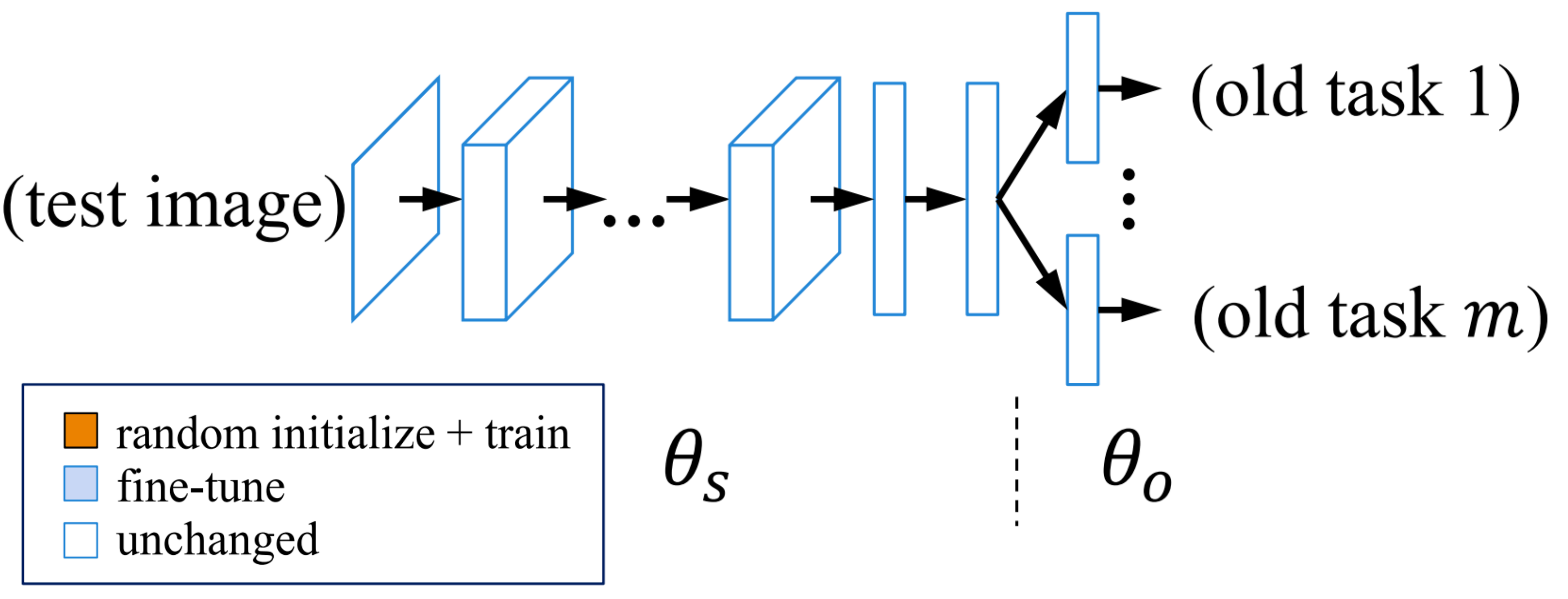}
  \caption{Teacher Model}\label{fig:lwf-teacher}
\end{subfigure}
\hfill
\begin{subfigure}{0.45\textwidth}
  \centering
  \includegraphics[width=\linewidth]{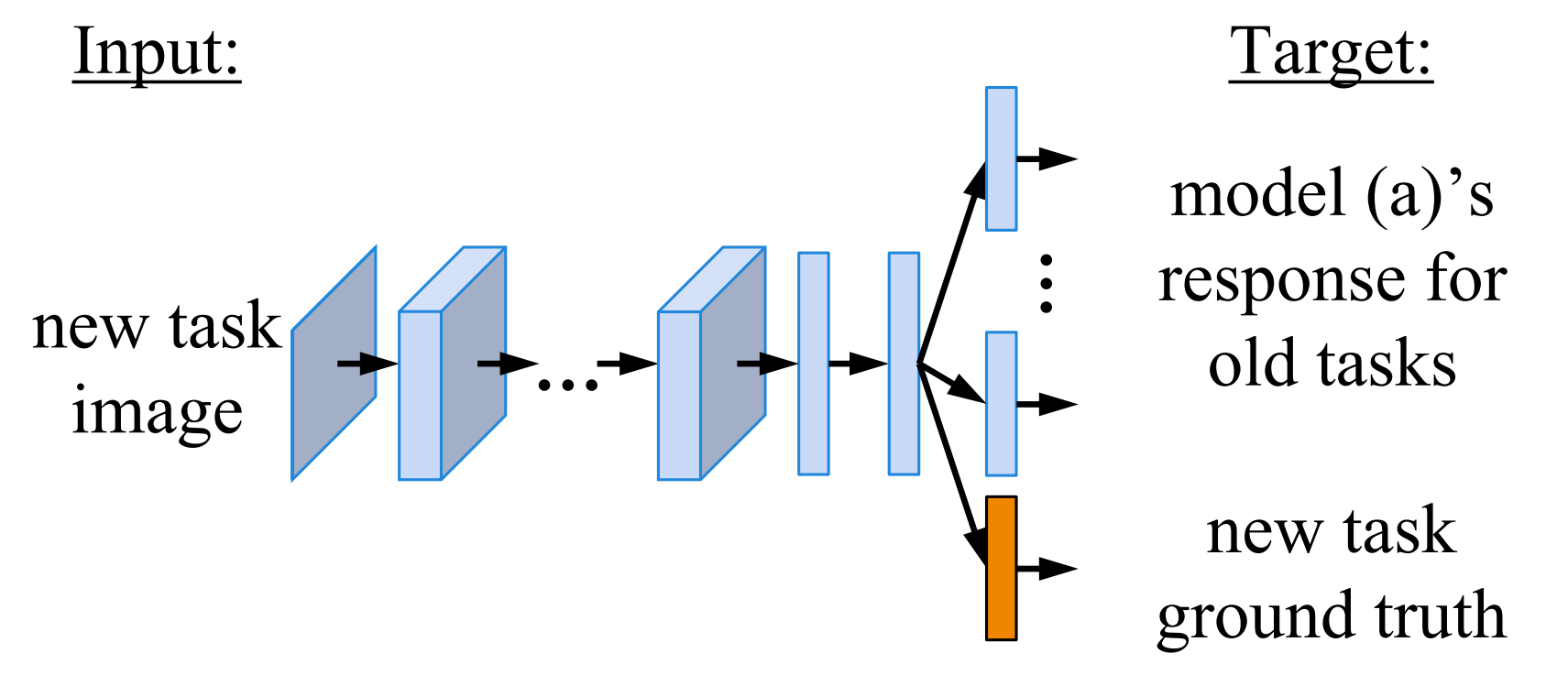}
  \caption{Student Model}\label{fig:lwf-student}
\end{subfigure}

\caption{The idea of Learning without Forgetting~\cite{li2017learningforgetting}}
\label{fig:lwfimage}

\end{figure*}

LwF is a functional regularization method that preserves the model’s behavior. Before the student model trains a new task, it saves a copy of itself as the teacher model, and then the student is required to match the teacher’s predictions via distillation loss to stabilize the student's output behavior~\cite{li2017learningforgetting}. As shwon in Figure~\ref{fig:lwf-teacher}, the structure contains shared parameters $\theta_s$, which extract features, and task-specific output parameters $\theta_o$, which are the predictions for learned tasks. After training on old tasks, these parameters are frozen as the teacher model. When learning a new task, shown in the Figure~\ref{fig:lwf-student}, $\theta_o$ for the new task is randomly initialized and trained using the ground-truth labels. At the same time, this model is constrained to reproduce the teacher model's outputs through distillation loss, which forces the student model to maintain the ability to handle the old tasks. Although LwF does not estimate parameter importance, these constraints indirectly restrict changes to parameters that strongly affect predictions on old tasks. Therefore, this method also has an implicit parameter regularization effect.

\subsection{Experience Replay}
\noindent Experience Replay is different from the above methods to prevent catastrophic forgetting, which maintains the performance of the model through dealing with the lack of information of old tasks. ER has a replay buffer that stores a small set of data samples from old tasks. When the model starts training on a new task, the training batch includes both the new task and the replay buffer. This allows gradient updates to perform well on the new task while not forgetting the old knowledge. This method helps stabilize the parameters and prevents representational drift in the hidden layers~\cite{lopezpaz2022gradientepisodicmemorycontinual}.
\subsection{Effective Rank (eRank)}
\noindent ERank is an effective way to track the expressive capacity and long-term health of the network. Recent work by Dohare et al. (Nature 2024) shows that structural collapse, where weight matrices become increasingly low rank, can be measured through eRank. As tasks accumulate, the singular values of feature subspaces become more concentrated~\cite{nature2024}. This leads to a sharp and lasting drop in eRank, reflecting the loss of plasticity. Then the network becomes less able to create new and independent feature directions, since the representational subspace narrows, and updates are forced into a small set of dominant directions. Because eRank captures these geometric changes directly, it provides a perspective for tracking how the structure of the model changes.

Several metrics have been proposed to explore and track forgetting or internal representations, including gradient and Fisher-based measures. However, these metrics are limited in CL settings. Gradient-based metrics, such as cosine similarity~\cite{lopezpaz2022gradientepisodicmemorycontinual}\cite{chaudhry2019efficientlifelonglearningagem}, quantify the alignment between gradients from the different tasks. The main idea is that forgetting occurs when gradient updates for new tasks conflict with the gradients that preserve old knowledge. These metrics identify the immediate causes of forgetting since gradients are typically computed at the batch or step level. Still, they fail to capture long-term loss of representational capacity and cannot explain situations in which the model cannot form new feature directions. Fisher-based metrics quantify the importance of each parameter for learned tasks, as parameters with high Fisher information are critical and should be protected during learning new tasks, such as Elastic Weight Consolidation~\cite{Kirkpatrick_2017}, Riemannian Walks~\cite{Chaudhry_2018} and Synaptic Intelligence~\cite{zenke2017continuallearningsynapticintelligence}. These measures explain forgetting from a parameter perspective but ignore the global structure of the learned representation space. Also, importance estimates accumulate noise as more tasks are learned, since they are computed repeatedly across different task distributions.

Unlike those metrics that focus on short-term behavior, eRank provides a global perspective on the internal structure. ERank can capture the situations in which the representation space collapses into a low-dimensional subspace. This allows tracking the loss of plasticity even when gradients do not conflict or when parameters are well regularized.

\subsubsection{Weight and Activation Effective Rank}
As described above, structural collapse is the degradation of the network’s complexity, and the weight matrix $W$ of a layer captures the capacity and complexity of the transformation that the layer can perform~\cite{nature2024}. eRank of $W$ measures the dimensionality of linear mapping. The low eRank indicates the structural collapse. The high eRank means the layer uses many structural degrees of freedom to transform its inputs. Tracking eRank of $W$ can monitor the loss of plasticity over time. A large drop across tasks shows that the network is losing its ability to create new transformations. This loss makes it harder to learn future tasks.

Except that structural collapse can be measured through eRank, representational collapse can also be measured. Representational collapse is the severe reduction in the dimensionality and diversity of the learned feature space in the hidden layers~\cite{saha2021spacestructuredcompressionsharing}, and the activation matrix $A$ is a feature covariance matrix. Thus, eRank of $A$ measures the dimensionality of the features that are sent to the next layer. When eRank is low, the features are compressed together, indicating the representational collapse. In this situation, the network loses the diverse representations it needs to separate tasks.
\subsection{Architectures}
\noindent In order to study the forgetting and collapse across different architectures, we include the following in this work.
\subsubsection{Multilayer perceptron (MLP)}
MLPs serve as a minimal baseline in continual learning. They have neither convolutional structure nor skip connections, so they are highly prone to catastrophic forgetting and representational collapse. They easily fall into sharp local minima~\cite{Rumelhart1986LearningRB} and rely heavily on large parameter updates to learn new tasks. These shifts speed up the overwriting of older representations. MLPs are also vulnerable to structural collapse. Their weight matrices lose rank quickly as tasks accumulate, which reduces the number of effective directions the network can use to learn. As this structural capacity shrinks, the model becomes rigid. This makes plasticity decline faster than in deeper or more structured architectures.
\subsubsection{ResNet-18}
ResNet-18 is a widely used convolutional architecture. It includes residual skip connections that improve optimization and keep the gradient flow stable~\cite{he2015deepresiduallearningimage}. These skip connections help preserve feature directions across tasks, which makes ResNet-18 perform better on catastrophic forgetting compared with MLPs. The network also has a deeper and more structured feature hierarchy~\cite{he2015deepresiduallearningimage}, which can slow down structural collapse since the feature space is spread across many layers. ResNet-18 is also less prone to representational collapse since its skip connections allow earlier features to pass forward without changes. This helps keep old feature directions active when learning new tasks. The multiple convolutional layers also create diverse and high-dimensional representations, which reduces the chance that features compress into a small set of dominant directions. Therefore, ResNet-18 is a more powerful baseline that deals with more complex tasks.
\subsubsection{Gated Recurrent Unit (GRU)}
Recurrent architecture extends representational capacity through adding temporal recurrence, which is that the model records an internal state that carries information from one timestep to the next~\cite{conv222}. The internal state is a short-term memory signal that updates with each new input, but does not truly store old knowledge. It only holds a small and temporary summary that fades over time.

ConvGRU combines convolution with gated recurrent units, which uses a sophisticated gating mechanism to regulate the flow of information~\cite{10.5555/2969239.2969329}. Thus, the gates can control how much past information is kept or replaced and allow the model to maintain useful old features while adding new ones. We use ConvGRU instead of pure recurrent architectures, such as GRU and LSTM, because pure recurrent layers operate on vectors, not feature maps. They remove spatial structure, which forces the model to flatten images or features and thereby reduces representational richness. ConvGRU keeps this through applying convolutions in the recurrent update.

Bi-ConvGRU is a bidirectional extension of ConvGRU. It contains two ConvGRU layers, one processes the sequence (the ordered series of feature maps) forward in time, while the other processes it backward~\cite{li2016videolstmconvolvesattendsflows}. The outputs of these two layers are combined afterward. This bidirectional extension allows the model to access both past and future context within a sequence, which stabilizes the learned features as information flows in two directions and reduces the chance that the model loses important signals. Therefore, the state updates with the traces of older information can slow the representational collapse and make the model more resistant to forgetting.
\section{Methodology}

\begin{algorithm}[!htb]
\caption{Training Procedure for Split MNIST (MLP, ConvGRU) and Split CIFAR-100 (ResNet-18, Bi-ConvGRU)}
\SetAlgoLined
\DontPrintSemicolon
\KwIn{
    Task sequence $\{T_1, \dots, T_K\}$; 
    Architecture $f_\theta$; 
    CL method $m \in \{\text{SGD}, \text{ER}, \text{LwF}\}$; 
    Replay buffer $\mathcal{B}$
}
\For{$t = 1$ \KwTo $K$}{
    \If{$m = \text{LwF}$}{
        Copy current model: $f_{\theta^*} \leftarrow f_\theta$ (frozen teacher)
    } 
    \ForEach{minibatch $(x, y)$ in $T_t$}{  
        \If{$m = \text{ER}$}{
            Sample $(x_B, y_B)$ from buffer $\mathcal{B}$ \\
            $x \leftarrow [x; x_B], \quad y \leftarrow [y; y_B]$
        }
        $\text{logits} \leftarrow f_\theta(x)$
        $\mathcal{L}_{\text{task}} = \text{CE}(\text{logits}, y)$
        \If{$m = \text{LwF}$}{
            $\text{logits}_{\text{teach}} = f_{\theta^*}(x) / T$ \\
            $\text{logits}_{\text{stud}} = \text{logits} / T$ \\
            $\mathcal{L}_{\text{distill}} = T^2 \cdot 
                \text{KL}(\text{softmax}(\text{logits}_{\text{teach}}) \;||\;
                              \text{softmax}(\text{logits}_{\text{stud}}))$ \\
            $\mathcal{L} \leftarrow \mathcal{L}_{\text{task}} + \lambda \cdot \mathcal{L}_{\text{distill}}$
        }
        \Else{
            $\mathcal{L} \leftarrow \mathcal{L}_{\text{task}}$
        }
        $\theta \leftarrow \theta - \eta \nabla_\theta \mathcal{L}$

        \If{$m = \text{ER}$}{
            Add $(x, y)$ into $\mathcal{B}$ via reservoir sampling
        }
    }
    Evaluate $f_\theta$ on tasks $\{T_1, \dots, T_t\}$ \\
}
\KwOut{
    Accuracy, Forgetting, Activation eRank, Grouped Weight eRank
}
\end{algorithm}

\noindent Our goal is to study the relationships between catastrophic forgetting and effective rank across multiple CL architectures. 
Therefore, the architectures are trained using a three-way comparison of CL strategies: SGD, ER, and LwF. Specifically, the MLP and ConvGRU are tested on Split MNIST, while the ResNet-18 and Bi-ConvGRU are tested on Split CIFAR-100. The training procedure for our experiment is listed in the Algorithm 1.
\subsection{Datasets}
\noindent The experiments use two datasets: Split MNIST and CIFAR-100. These two datasets are different in complexity and difficulty. Split MNIST provides a simple environment while Split CIFAR-100 provides a high-dimensional and more challenging environment.
\subsubsection{Split MNIST (Task-Incremental Learning)}
The original MNIST dataset, containing 60,000 training images and 10,000 test images, is divided into five tasks under the Task-IL setting. As shown in the Figure~\ref{fig:splitmnist}, ten MNIST digits are separated into five binary classification tasks. Each of the five boxes in the image represents a single task that contains a pair of previously unseen digits. For example, Task 1 contains only the digits ‘0’ and ‘1’, Task 2 contains only ‘2’ and ‘3’, and so on. In order to train each individual binary task, the original digit labels (0, 1, …, 9) are remapped to the local class label \{{0,1\} for each task, thus each task becomes mutually disjoint. This design is suitable for observing the changes in eRank and analyzing the structural and representational collapse directly.

The dataset is preprocessed identically for each model. Each pixel of the image is normalized to roughly zero mean and unit variance. For the input of MLP, the images are flattened into vectors, and ConvGRU uses the original image shape ($1\times28\times28$).

\begin{figure*}[t]
\centering
\includegraphics[width=\textwidth]{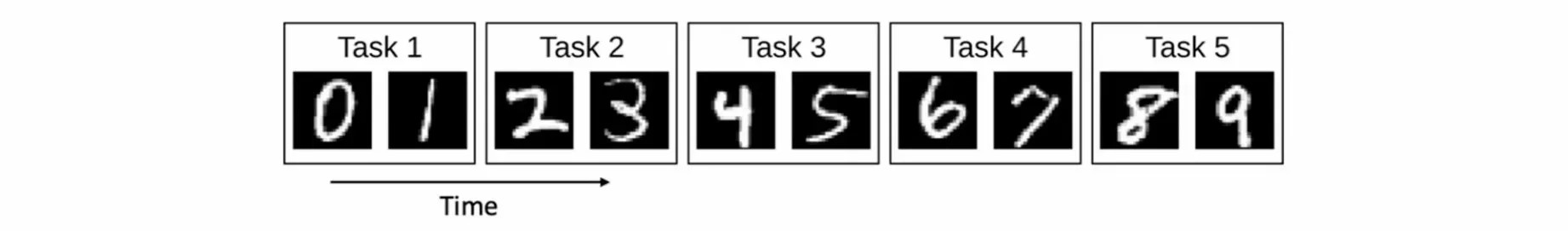}
\caption{Five sequential binary classification tasks of Split Mnist dataset}\label{fig:splitmnist}
\end{figure*}

\subsubsection{Split CIFAR-100 (Class-Incremental Learning)}
The original CIFAR-100 dataset contains 50,000 training images and 10,000 test images across 100 classes. This dataset is divided into 20 tasks under Class-IL, and each task contains five separate classes. Thus, the model uses one shared softmax head for all seen classes. The task identity is not provided during the evaluation, so the model needs to classify the images from all previous tasks. The task boundaries exist under this setting, which means that the model starts training on a new task without seeing the past tasks. This is useful to define when forgetting and collapse begin.

The images in the dataset are converted to tensors in size $3\times32\times32$. They are normalized using the mean and standard deviation computed from the entire dataset. 
\subsection{Architectures}
\noindent To match the models to the dataset's complexity, we evaluate four architectures across two benchmarks. For the simpler Split MNIST, we use a simple MLP and a convolutional GRU to handle the simple digits. For the more challenging Split CIFAR-100 dataset, we use a standard ResNet-18 and a bidirectional ConvGRU. These two architectures have higher capacity to handle the greater visual complexity and richer class structure.
\subsubsection{Mltilayer Perceptron (MLP)}
MLP acts as a baseline architecture because of the absence of the convolutional structure, which is highly prone to forgetting and structural collapse. It allows us to observe rapid structural and representational collapse and define the lower bound of robustness for our study. Our MLP contains two fully connected hidden layers, an input layer, and an output layer. The activation function is ReLU for hidden layers.  The calculation for the network is simply moving forward:
\[
    x \to \text{ReLU}(W_1x + b_1) \to \text{ReLU}(W_2h + b_2) \to \text{FC}_{10}(h),
\]
where $W$ for weight matrix and $b$ for bias vector. The Stochastic Gradient Descent (SGD) is used. The learning rate is 0.01, and the momentum is 0.9. This simple feedforward network is prone to forgetting and collapse, which can clearly observe the representational drift and degradation of $eRank$. 
\subsubsection{Convolutional Gated Recurrent Unit (ConvGRU)}
ConvGRU is the intermediate architecture between MLP and ResNet-18 in our study. It is a hybrid network that incorporates gated recurrence to the convolutional feature maps. This action allows the network to preserve the old representations more effectively, and helps to investigate whether temporal memory and gated information flow help preserve representational dimensionality. From the eRank perspective, this network also helps to study how recurrence affects both structural and representational collapse compared to purely feedforward networks. Because of the simplicity of the Split MNIST, the ConvGRU has three stages: a convolutional feature extractor, a gated recurrent unit, and a classification output. The network begins with four convolutional layers to extract the spatial feature:
\[
\begin{aligned}
h_1 &= \sigma(\mathrm{Conv}_1(x)), \\
h_2 &= \sigma(\mathrm{Conv}_2(h_1)), \\
h_3 &= \sigma(\mathrm{Conv}_3(h_2)), \\
h_4 &= \sigma(\mathrm{Conv}_4(h_3)).
\end{aligned}
\]

\noindent Each convolution layer uses $3\times3$ kernel and ReLU activation, and the output of the final convolutional layer $h_4$ is passed to the recurrent block. 

GRU helps the network memorize important information during data processing. This part is implemented using convolutional operations, which allow it to update the hidden state $h_t$ across spatial feature maps~\cite{conv222}. The $h_t$ in our work is a 3D tensor because the network's input is images. Two gates and one candidate state decide the update of $h_t$. Figure~\ref{fig:convgru} shows the main structure of our GRU. The reset gate $r_t$, which acts as a filter, analyzes which parts of the previous state are relevant to the current features. In our implementation, the current input $x_t$ and the previous state $h_{t-1}$ are processed by convolutional kernels, so the reset gates $r_t$ are spatial masks, similar to the update gate later. If $r_t$ of a pixel is 0, then it suggests that the past information at this location should be reset. The equation is:
\[r_t = \sigma\left(W_r * x_t + U_r * h_{t-1} + b_r \right).\]

\begin{figure*}[t]
\centering
\includegraphics[width=0.5\textwidth]{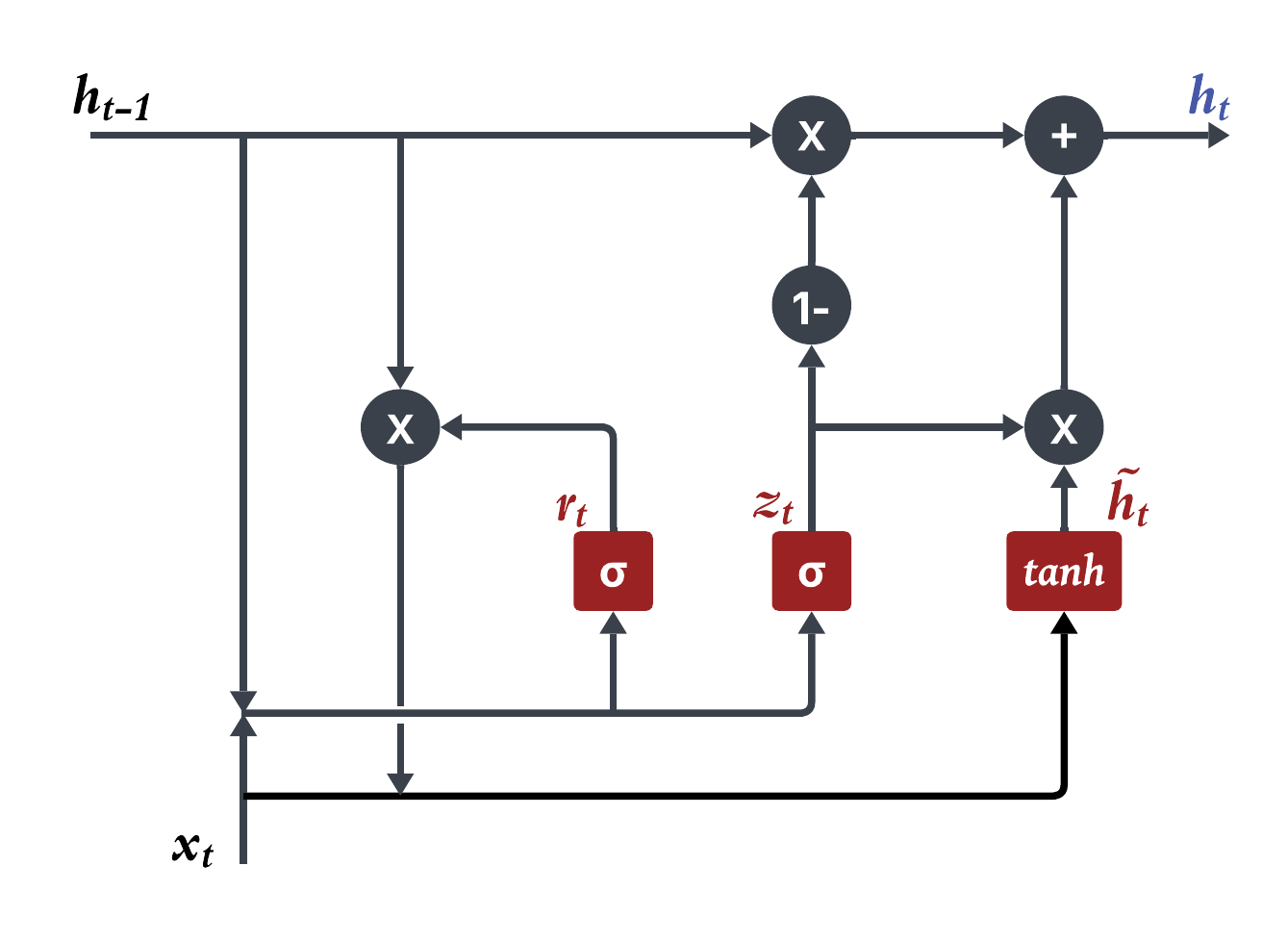}
\caption{The update of hidden state $h_t$ is defined by Update Gate $z_t$, Reset Gate $r_t$ and Candidate State $\tilde{h}_t$ in Gated Recurrent Block of ConvGRU}\label{fig:convgru}
\end{figure*}

\begin{figure*}[t]
\centering
\includegraphics[width=\textwidth]{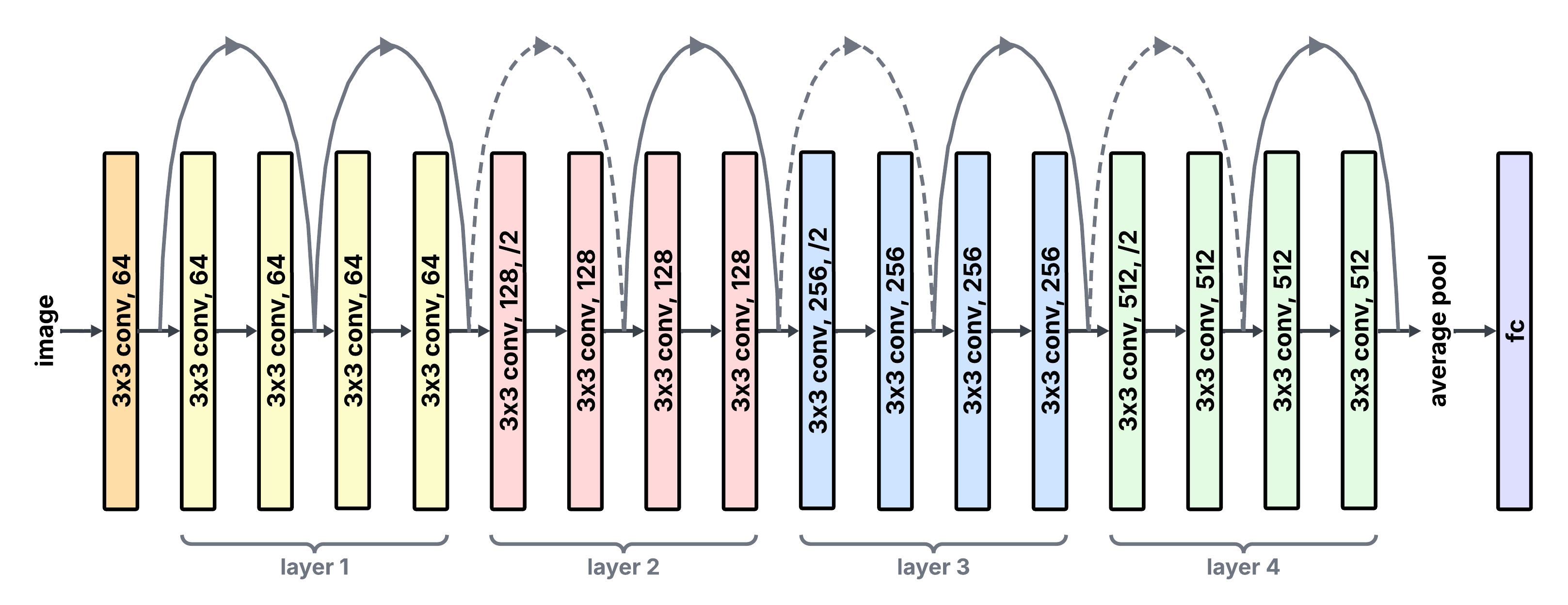}
\caption{The Architecture of ResNet-18}\label{fig:resnet18}
\end{figure*}

\noindent The update gate $z_t$ decides the amount of the previous state $h_{t-1}$ and the new candidate state $\tilde{h}_t$ to pass forward:
\[z_t = \sigma\left(W_z * x_t + U_z * h_{t-1} + b_z \right).\]
If $z_t$ is close to 1 of a specific pixel, it keeps the old state and forgets the new input. If it is close to 0, it overwrites the old state. Then, the candidate state $\tilde{h}_t$ represents the potential new state but has not yet been updated. It contains all the new extracted features and the decision made from the reset gates:
\[\tilde{h}_t = \tanh\left(W_h * x_t + U_h * (r_t \odot h_{t-1}) + b_h \right).\] 

\noindent Finally, the recurrent output is calculated through:
\[h_t = (1 - z_t) \odot h_{t-1} + z_t \odot \tilde{h}_t.\]

In the classification process, the output of the recurrent block ($h_t$) is passed through the average pooling layer to flatten the feature map into a vector, and then map to the class logits.

Overall, this network includes convolutional spatial structure and recurrent gates that are absent in the MLP. These changes ensure the representations change more smoothly to achieve the purpose of slowing down collapse and to investigate the correlation between $eRank$ collapse and forgetting.

\subsubsection{ResNet-18}
ResNet-18 is selected because many previous works have already demonstrated robustness and good performance on the CIFAR-100 dataset~\cite{he2015deepresiduallearningimage}~\cite{pfülb2019comprehensiveapplicationorientedstudycatastrophic}. Especially, the skip connections used in the network directly affect the stability of the representation. It also allows us to investigate how architectural mechanisms that improve gradient flow and feature reuse affect representational collapse. Figure~\ref{fig:resnet18} shows the architecture we used, which includes an initial convolutional stage (first block in the figure), a residual learning stage (layers 1 to 4), a global average pooling layer, and a fully connected classification layer. In residual stages, each layer contains two residual blocks, which stabilize the flow of gradients to reduce representational collapse.
\subsubsection{Bi-ConvGRU}
Compared with ConvGRU, Bi-ConvGRU uses bidirectional recurrence to better handle the Split CIFAR-100. This change allows the network to learn two complementary spatial memories, which outputs the richer final representation to reduce collapse. In the Split CIFAR-100 setting, it allows us to test whether stronger temporal integration can further preserve representational dimensionality and reduce forgetting in a more complex environment. Instead of time flowing one way: $t \to t+1$, the network uses two GRU layers that operate in parallel but process the same feature map in opposite directions. Each layer has its own set of weights. The forward layer understands history (from “cause” to “effect”):
\[h_t^{(f)} = \text{GRU}(x_t, h_{t-1}^{(f)}),\]
and the backward layer understands the future context (from “effect” to “cause”):
\[h_t^{(b)} = \text{GRU}(x_t, h_{t+1}^{(b)}).\]
Then, the final output is a combination of what the forward layer remembers and what the backward layer anticipates:
\[h_t^{(bi)} = [h_t^{(f)} || h_t^{(b)}] .\]
This output doubles the channel dimension and expands the recurrent feature space.

\subsection{Continual Learning Strategies}
\noindent Three continual learning strategies are used, including vanilla SGD, ER, and LwF, which represent three approaches to reducing catastrophic forgetting. The SGD is a baseline that does not provide a specific strategy to prevent forgetting, which shows the full extent of forgetting of the network. ER encourages the reuse of feature directions from past tasks to improve both structural and representational collapse. LwF restricts changes in critical parameters to reduce forgetting, but it may still exhibit representational collapse in shared feature space. Comparing these strategies allows us to observe how replay and output constraints affect the preservation of internal structure and representational capacity in CL settings.
\subsubsection{Experience Replay}
In both experiments, the same ER framework is used. The replay buffer stores input examples and their corresponding labels by global class indexing, where 0-9 for MNIST and 0-99 for CIFAR-100. The buffer has a fixed capacity with a maximum of 2000 samples. The reservoir sampling is used to ensure that all samples have an equal probability of being retained in the buffer. 

At each training step of tasks, a minibatch is sampled from the current tasks, and then an equal-sized batch is sampled randomly from the replay buffer. These two batches are concatenated. Finally, the model performs an SGD update to minimize the loss on the new task and the replayed data.
\subsubsection{Learning without Forgetting}
Our LwF implementation follows the classic setup. Before training on the new task, a copy of the current model is made and frozen to serve as the teacher network ($\theta^{(\text{teacher})})$. During training, functional regularization is implemented to preserve output behavior on previous tasks while learning new:
\[\mathcal{L} = \mathcal{L}_{\text{task}} + \lambda \cdot \mathcal{L}_{\text{distill}},\]
where $\mathcal{L}_{\text{task}}$ is the task loss and $\mathcal{L}_{\text{distill}}$ is 
the distillation loss. $\mathcal{L}_{\text{distill}}$, which encourages the model to maintain previous outputs, is calculated by the KL divergence between the soft probability distributions of the teacher and current (student) models:
\[\mathcal{L}_{\text{distill}} = T^2 \cdot \mathrm{KL}\left( p^{(\text{teacher})} \;\|\; p^{(\text{student})} \right),\]
where $T$ is the temperature scaling. Soft probability is used when training the student to match the teacher because this allows the student to learn relative decision boundaries more easily and efficiently than trying to match abrupt 0s and 1s, which leads the rapid weights changes and unstable training~\cite{li2017learningforgetting}. If $T> 1$, the distribution becomes softer. When tasks are more complicated, a higher $T$ is used. We use $T = 2$ in the experiments.

\subsection{Metrics}
\noindent We include 4 metrics to quantify the model performance and internal structural behavior, which include the average accuracy, forgetting, effectvie rank, and normalized metric for cross-architecture comparison.
\subsubsection{Accuracy and Forgetting}
After training on task $t$, the task accuracy is calculated:
\[
\mathrm{Acc}(\tau \mid t) =
\frac{1}{|\mathcal{D}_\tau|}
\sum_{(x,y)\in\mathcal{D}_\tau}
\mathbf{1}[\hat{y}=y].
\]
Then the average accuracy aggregates the performance of all previous tasks:
\[A = \frac{1}{t} \sum_{\tau=1}^{t} \text{Acc}(\tau \mid t).\]
The forgetting metric measures the degradation in performance. For a task $k$, the forgetting is defined as:
\[F_k = \max_{t < T} \text{Acc}(k \mid t) - \text{Acc}(k \mid T).\]
\subsubsection{Effective Rank}
The eRank quantifies the dimensionality of a matrix using the entropy of singular values. Given a matrix with singular values $\sigma_i$:
\[\text{eRank} = \exp\left( -\sum_i p_i \log p_i \right),\]
where $p_i$ is the normalized singular value distribution. Activation eRank is calculated on the hidden representations $H$ of the penultimate layer, while weight eRank is computed for the weight matrices of each layer.

The different architectures have different numbers of layers and parameter shapes, so it is meaningless to compare the weight eRank of each layer. Thus, we group the layers into comparable blocks as listed in Table~\ref{tab:layer_grouping}.

\begin{table}[]
    \centering
    \begin{tabular}{@{}lcccc@{}}
        \toprule
        \textbf{Group} & \textbf{MLP} & \textbf{ConvGRU} & \textbf{ResNet-18} & \textbf{Bi-ConvGRU} \\
        \midrule
        \textbf{Early} & FC1 & Conv Layers & Residual Stage 1 & Conv Layer 1 \\
        \textbf{Middle} & FC2 & GRU & Residual Stage 2-3 & Conv Layers 2-4\\
        \textbf{Late} & --- & --- & Residual Stage 4 & GRU \\
        \textbf{Head} & Classifier & Classifier & Classifier & Classifier \\
        \bottomrule
    \end{tabular}
    \caption{Layer Grouping for Weight eRank}
    \label{tab:layer_grouping}
\end{table}

The peak-normalized eRank ($\text{eRank}_{pct}$) is used, since different architectures have different layer dimensions. The $\text{eRank}_{pct}$ is the percentage of the best historical eRank that is still retained in the training step: 
\[\text{eRank}_{pct} = \frac{\text{eRank}(t)}{\max_{u \le t} \text{eRank}(u)}.\]

\section{Experimental Results}

\begin{figure*}[t]
\centering

\begin{subfigure}{0.48\linewidth}
\centering
\includegraphics[width=\linewidth]{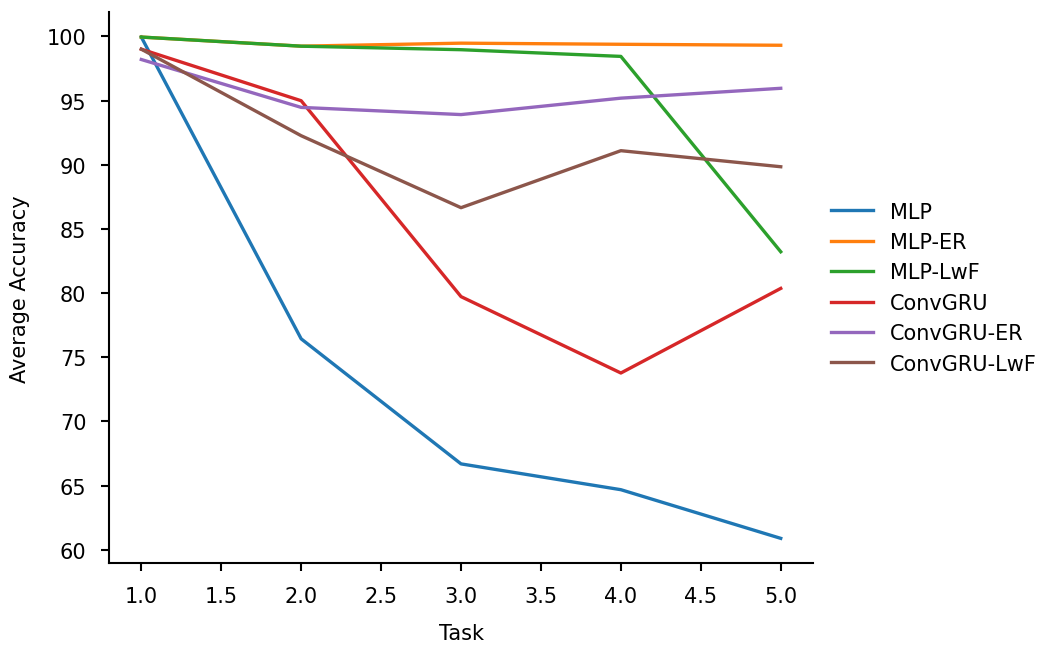}
\caption{Average accuracy for Split MNIST}
\label{fig:mnist-acc}
\end{subfigure}
\hfill
\begin{subfigure}{0.48\linewidth}
\centering
\includegraphics[width=\linewidth]{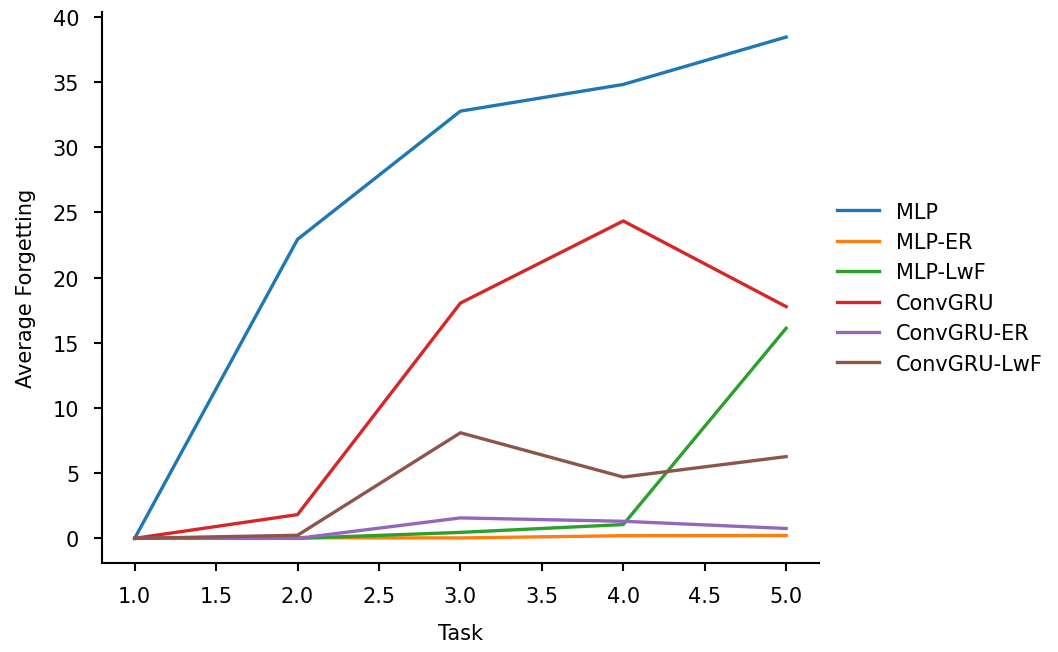}
\caption{Forgetting curve for Split MNIST}
\label{fig:mnist-forgetting}
\end{subfigure}

\caption{SGD (the pure network) shows severe forgetting, particularly for MLPs. LwF performs better but still has late-task degradation. ER is most effective across all tasks.}
\label{fig:mnist-acc+forgetting}

\end{figure*}

\begin{figure}[t]
\centering
\includegraphics[width=0.48\textwidth]{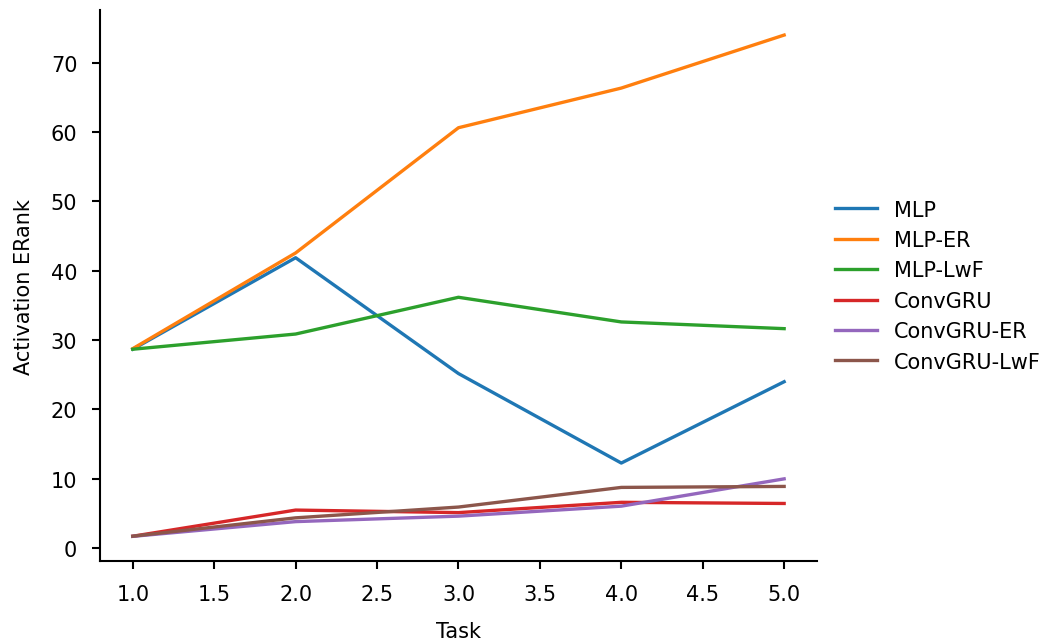}

\caption{Activation eRank For Split MNIST - Activation eRank collapses for pure MLP, stabilizes under MLP-LwF, and increases under MLP-ER, while ConvGRU maintains low eRank across all strategies due to recurrent gating.}
\label{fig:mnist-acterank}

\end{figure}

This section demonstrates the empirical findings derived from the Split MNIST and Split CIFAR-100 experiments under different architectures and CL strategies. The results are visualized and analyzed using performance metrics and structural metrics.

Figures~\ref{fig:mnist-acc} and~\ref{fig:mnist-forgetting} present the average accuracy and forgetting curves tested on the Split MNIST. The results show a clear degradation under the SGD strategy, where both MLP and ConvGRU experience a sharp decline in accuracy and forgetting over time. MLP suffers the strongest decline, where forgetting starts in Task 2 and continues to fail to 40\% in Task 5. ConvGRU performs better, but still suffers from forgetting under SGD. The application of LwF shows noticeable improvement in both architectures. The accuracy and forgetting curves stabilize in later tasks compared to SGD. MLP-LwF experiences almost no forgetting in the first four tasks, but a sudden decline in the fifth, while MLP-LwF suffers more forgetting but has stable performance over all tasks. The ER strategy shows the greatest improvement. Both models maintain high accuracy and nearly no forgetting over all tasks. MLP-ER performs slightly better than ConvGRU-ER.

Figure~\ref{fig:mnist-acterank} illustrates how activation eRank changes across tasks for different architectures and learning strategies. For MLP under SGD, activation eRank increases early but collapses as more tasks are learned, which falls from roughly 45 at task 2 to roughly 12 at task 4. This decline is consistent with the sharp increase in forgetting observed previously. For MLP-LwF, activation eRank is similar during the whole process. In contrast, MLP-ER continues to increase and reaches the highest eRank in the final task. ConvGRU shows consistently low activation eRank regardless of the training strategies, since recurrent gating repeatedly compresses representations into a small subspace, which will be explained further through peak-normalized activation eRank in the Discussion section. 

To observe structural changes in feature extractors, we first explore the weight eRank of the early layers in the MLP (Figure~\ref{fig:mnist-werank-mlp}). The eRank shows a clear decline, which indicates a gradual structural compression of the first hidden. In contrast, MLP-LwF slightly underperforms the baseline SGD, indicating that functional regularization does not protect the early weights from structural collapse. MLP-ER stays consistently above the SGD curve and decays more slowly, suggesting that ER helps preserve richer features. A similar pattern appears in ConvGRU (Figure~\ref{fig:mnist-werank-gru}). As shown in Figures~\ref{fig:mnist-ermlp} and~\ref{fig:mnist-ergru}, the eRank trajectories in the middle-layer are similar to the early-layer curves. This shows that structural collapse is not limited to the early stage, but also affects deeper layers. Figure~\ref{fig:mnist-werhead} shows the weight eRank for the classification layer. All models experience a sharp decline on Task 2 since learning new classes requires reusing and overwriting previous decision boundaries. ER and LwF reduce this effect but cannot eliminate it for both architectures. For MLPs, ER maintains the highest eRank throughout training, suggesting that stored exemplars help preserve the discriminative subspace for earlier classes. LwF improves, but remains closer to SGD. For ConvGRUs, the eRanks are similar for different strategies, but ER and LwF perform a more stable curve. In general, these plots show that the models undergo structural collapse, and ER consistently slows this process, whereas LwF shows limited improvement.
\begin{figure*}[!t]
\centering

\begin{subfigure}{0.48\linewidth}
\centering
\includegraphics[width=\linewidth]{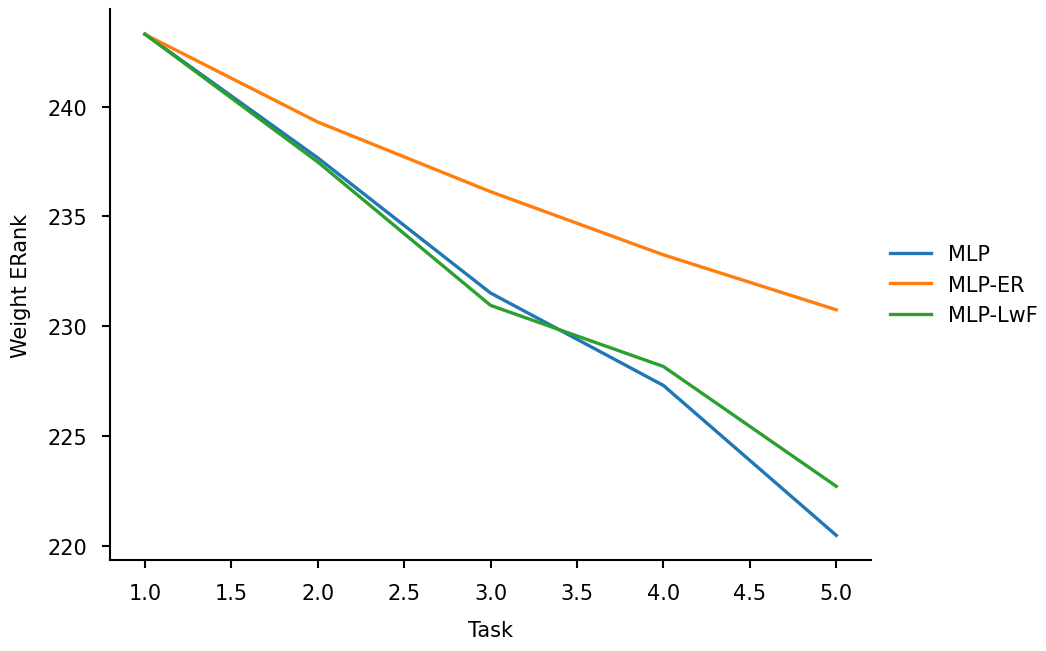}
\caption{Weight eRank of MLP Variants in Early Layers For Split MNIST}\label{fig:mnist-werank-mlp}
\end{subfigure}
\hfill
\begin{subfigure}{0.48\linewidth}
\centering
\includegraphics[width=\linewidth]{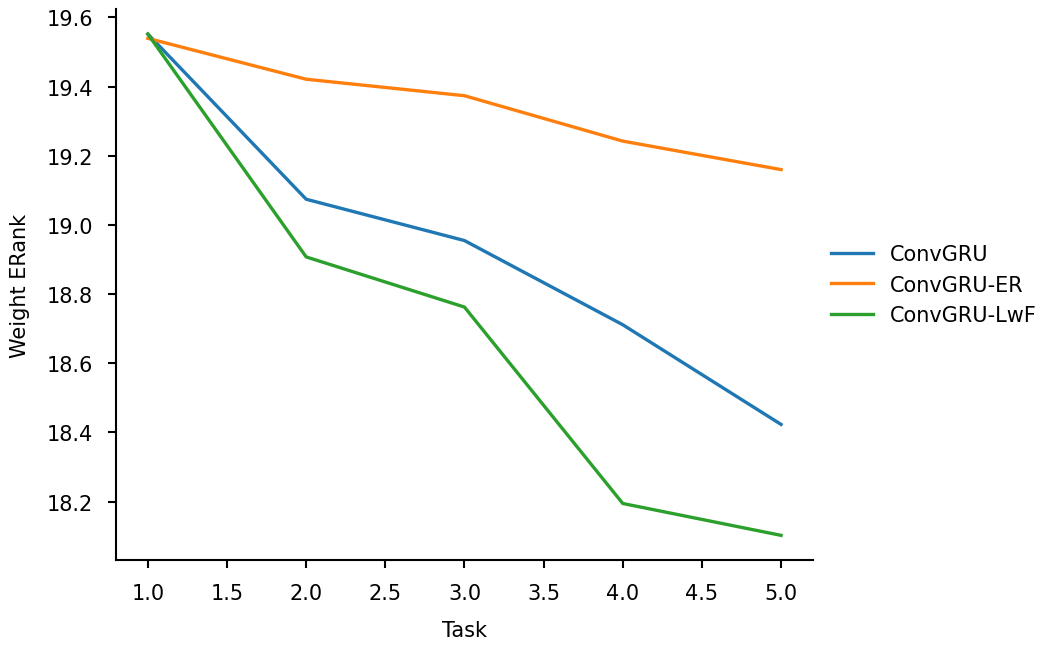}
\caption{Weight eRank of ConvGRU Variants in Early Layers For Split MNIST}\label{fig:mnist-werank-gru}
\end{subfigure}

\vspace{0.3cm}

\begin{subfigure}{0.48\linewidth}
\centering
\includegraphics[width=\linewidth]{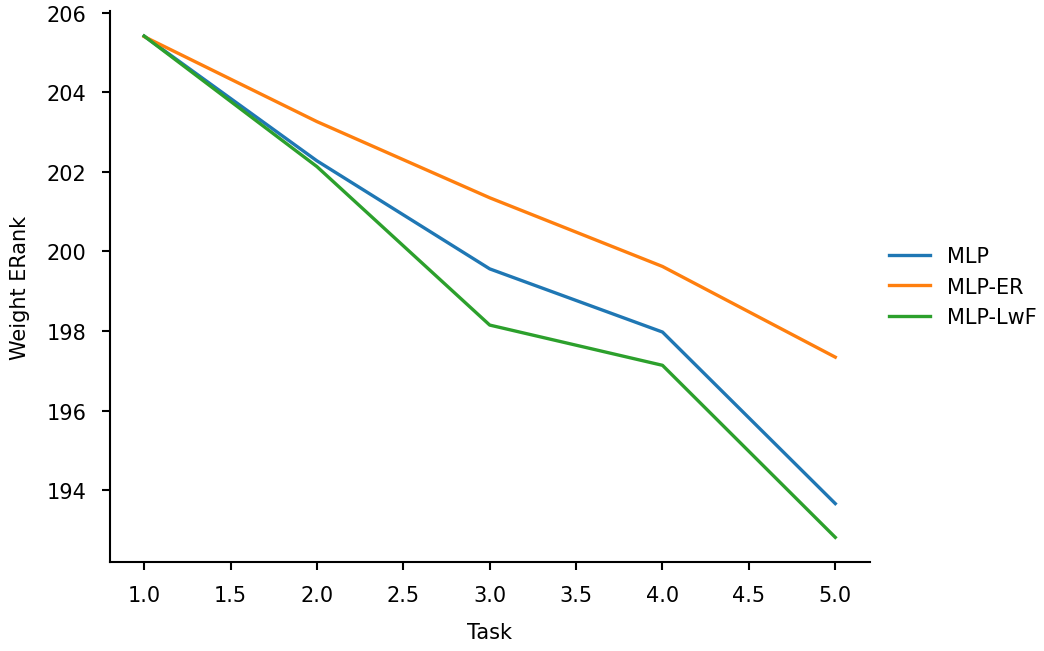}
\caption{Weight eRank of MLP Variants in Middle layers For Split MNIST}\label{fig:mnist-ermlp}
\end{subfigure}
\hfill
\begin{subfigure}{0.48\linewidth}
\centering
\includegraphics[width=\linewidth]{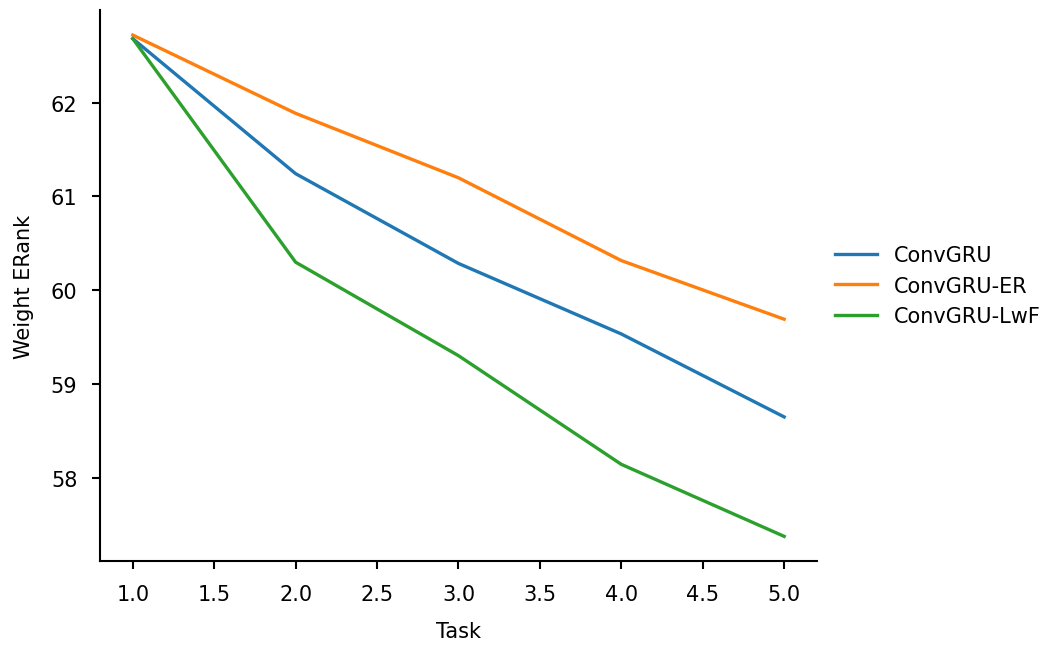}
\caption{Weight eRank of ConvGRU Variants in Middle Layers For Split MNIST}\label{fig:mnist-ergru}
\end{subfigure}

\vspace{0.3cm}

\begin{subfigure}{0.48\linewidth}
\centering
\includegraphics[width=\linewidth]{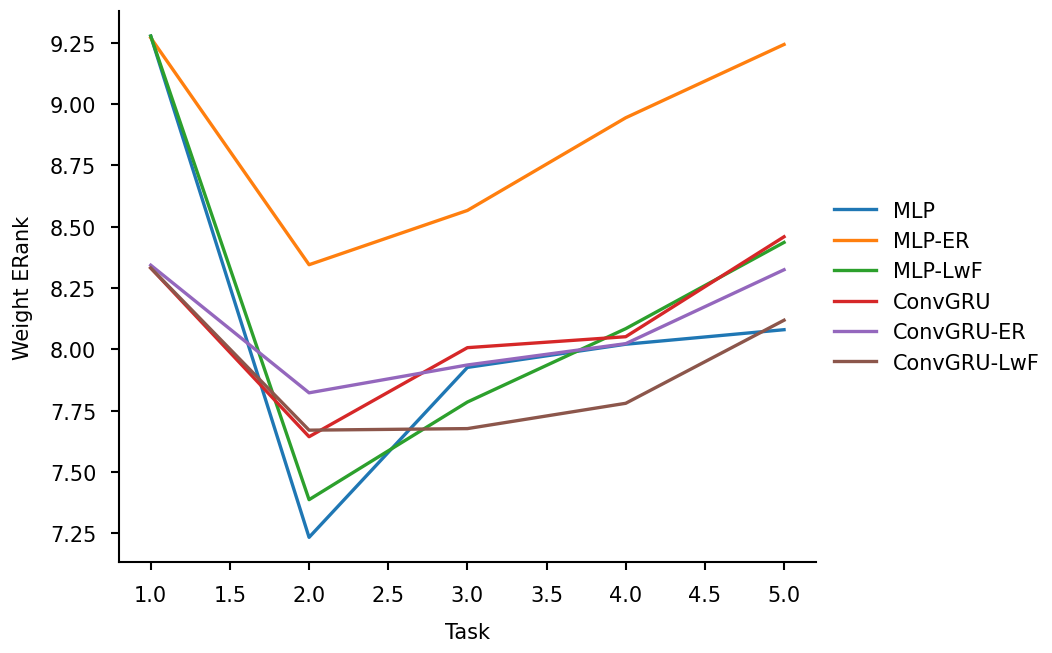}
\caption{Weight eRank of in Classification Layers For Split MNIST}\label{fig:mnist-werhead}
\end{subfigure}

\caption{Weight eRank indicates structural collapse across all layers in both architectures. Apply ER clearly slows the degradation, whereas LwF yields only limited improvement, particularly in the early and middle layers.}
\label{fig:mnist-werank}

\end{figure*}

Figure~\ref{fig:cifar-acc} presents the average accuracy in 20 tasks for ResNet-18s and Bi-ConvGRUs. Both SGD-trained architectures show a clear degradation, with a dramatic drop in accuracy after Task 1 and only about 20\% in later tasks. In contrast, LwF performs better. In the late stage of training, ResNet-18 has an accuracy of nearly 60\%, while Bi-ConvGRU has a lower performance with 40\% accuracy. ER shows the most significant improvements, where both models maintain nearly 80\% accuracy throughout the training. The forgetting curves, shown in Figure~\ref{fig:cifar-forgetting}, are consistent with the accuracy curves.

Figure~\ref{fig:cifar-acter} shows the activation eRank across tasks. Compared to Bi-ConvGRU, ResNet-18 has a higher eRank in early tasks due to its higher initial activation dimensionality. This helps the model to have strong representational richness in early tasks. However, under the SGD strategy, its eRank eventually collapses to nearly zero. ER and LwF show significant improvement and maintain the higher eRank value throughout training. For Bi-ConvGRUs, ER and LwF also stabilize eRank, whereas LwF indicates only a slight improvement. 

\begin{figure*}[t]
\centering

\begin{subfigure}{0.48\linewidth}
\centering
\includegraphics[width=\linewidth]{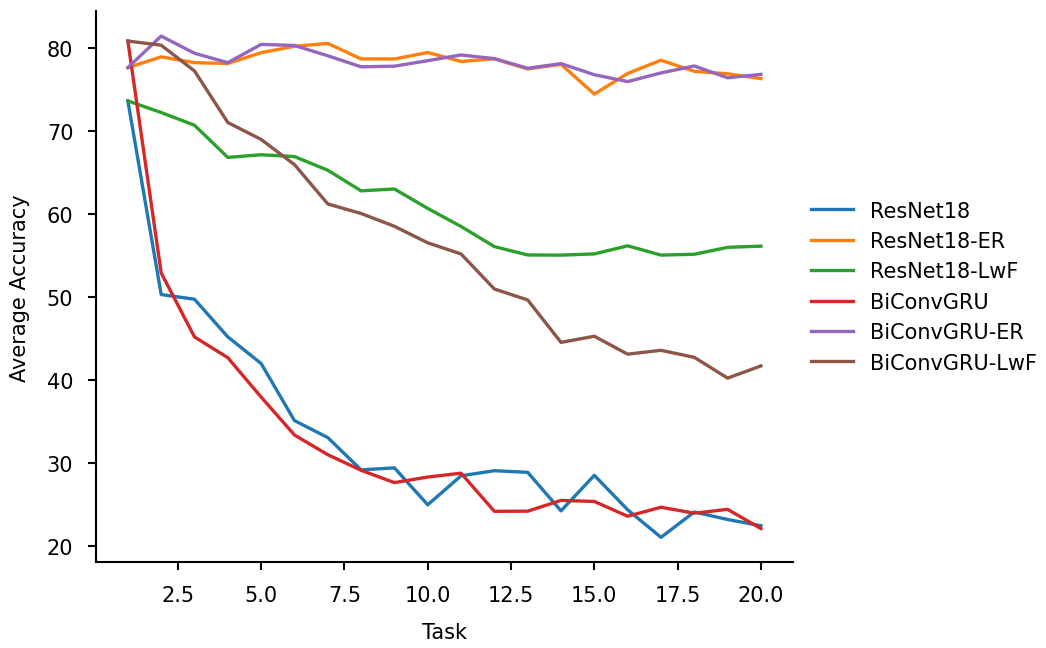}
  \caption{The Average Accuracy Curve For Split CIFAR-100}\label{fig:cifar-acc}
\end{subfigure}
\hfill
\begin{subfigure}{0.48\linewidth}
\centering
\includegraphics[width=\linewidth]{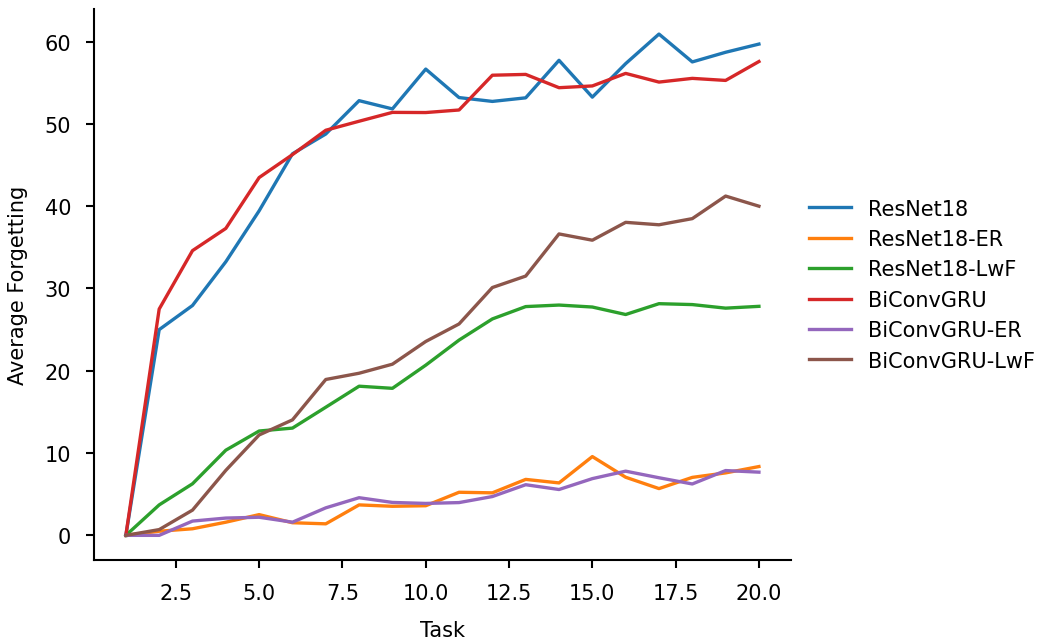}
\caption{The Forgetting Curve For Split CIFAR-100}\label{fig:cifar-forgetting}
\end{subfigure}

\caption{Both ResNet-18 and Bi-ConvGRU suffer severe accuracy degradation under SGD. LwF mitigates forgetting, and ER achieves the best performance.}
\label{fig:cifar-acc+forgetting}

\end{figure*}

\begin{figure}[t]
\centering
\includegraphics[width=0.48\textwidth]{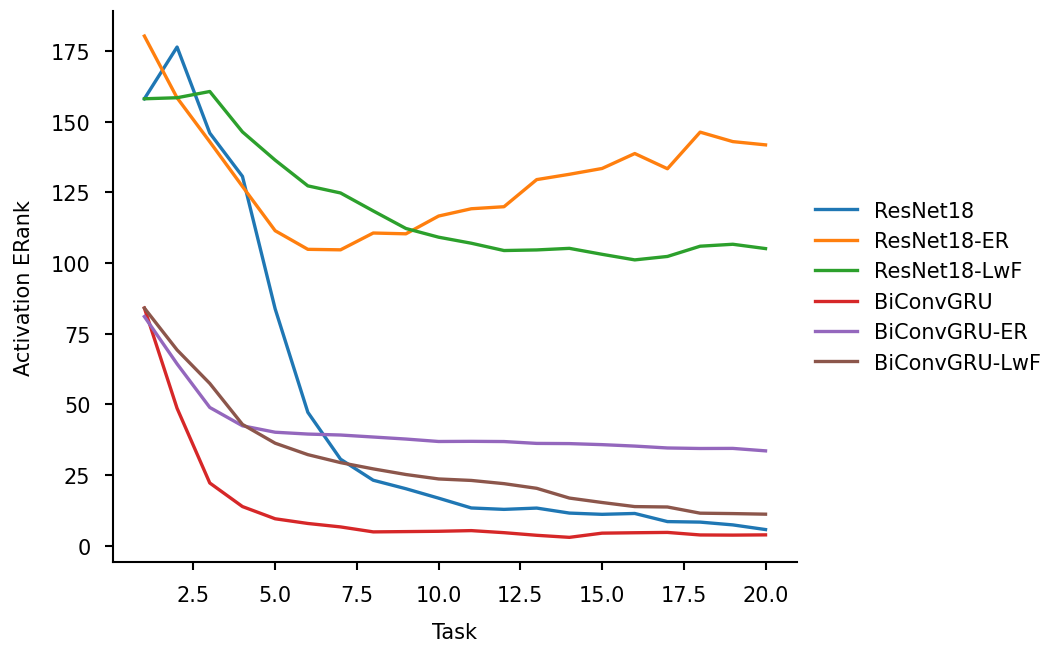}

\caption{The Activation eRank Curve For Split CIFAR-100 - Activation eRank collapses severely for each pure architectures. ER and LwF both improve representational collapse, but ER outperforms LwF.}
\label{fig:cifar-acter}

\end{figure}

Unlike in Split MNIST, the early and middle layers of Bi-ConvGRU are convolutional layers, so that they can be directly compared without peak normalization. As shown in Figure~\ref{fig:cifar-wer-1}, all models start with an weight eRank of 60, which shows the initial diversity and complexity of the kernels that extract features. As tasks accumulate, all methods show an initial sharp decline in the first five tasks since the structural collapse occurs when the network optimizes for the shared feature space. ResNet-18-SGD shows a continuous decline, which is worse than Bi-ConvGRU. This indicates that skip connections are less effective in preventing the structural and long-term weight collapse in the early layers compared to the gated memory of the Bi-ConvGRU. For ResNet-18, the LwF strategy suffers the most severe collapse among all models, whereas it shows a slight improvement in Bi-ConvGRU. ER is more effective for both architectures. The middle layers have higher start eRank values than the early layers (Figure~\ref{fig:cifar-wer-2}), which indicates the complex and higher-dimensional feature maps in these layers. Under SGD, Bi-ConvGRU suffers the most collapse among all models, while ResNet-18 performs better due to skip connections. The ER strategy helps both architectures stabilize eRank at a high value, while LwF is less effective at stabilizing to preserve the full dimensionality of the internal weights required for this task. 

Figure~\ref{fig:cifar-wer-3} shows the changes in the weight eRank in the late layers, where ER and LwF show similar patterns as in the middle layers. ResNet-18 shows an apparent decrease and drops below 50 in the final task. This indicates that most representational dimensions become unused and make the layer unable to learn new features. Bi-ConvGRU also shows a significant collapse that falling from near 180 to 70. The weight eRank curves in Figure~\ref{fig:cifar-wer-head} show how the classification layers change. Both architectures with ER demonstrate a recovery in curves, especially ResNet-18, which shows an eventual rise. All other models stabilize at low eRank values.

\begin{figure*}[!t]
\centering

\begin{subfigure}{0.48\linewidth}
\centering
\includegraphics[width=\linewidth]{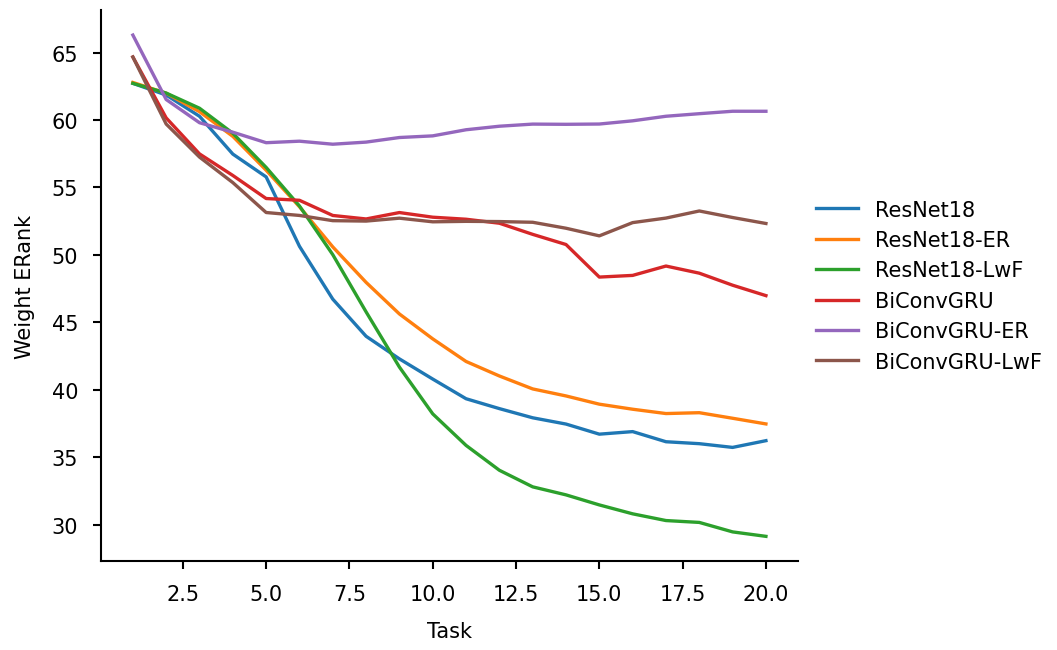}
\caption{The Weight eRank in Early Layers For Split CIFAR-100}\label{fig:cifar-wer-1}
\end{subfigure}
\hfill
\begin{subfigure}{0.48\linewidth}
\centering
\includegraphics[width=\linewidth]{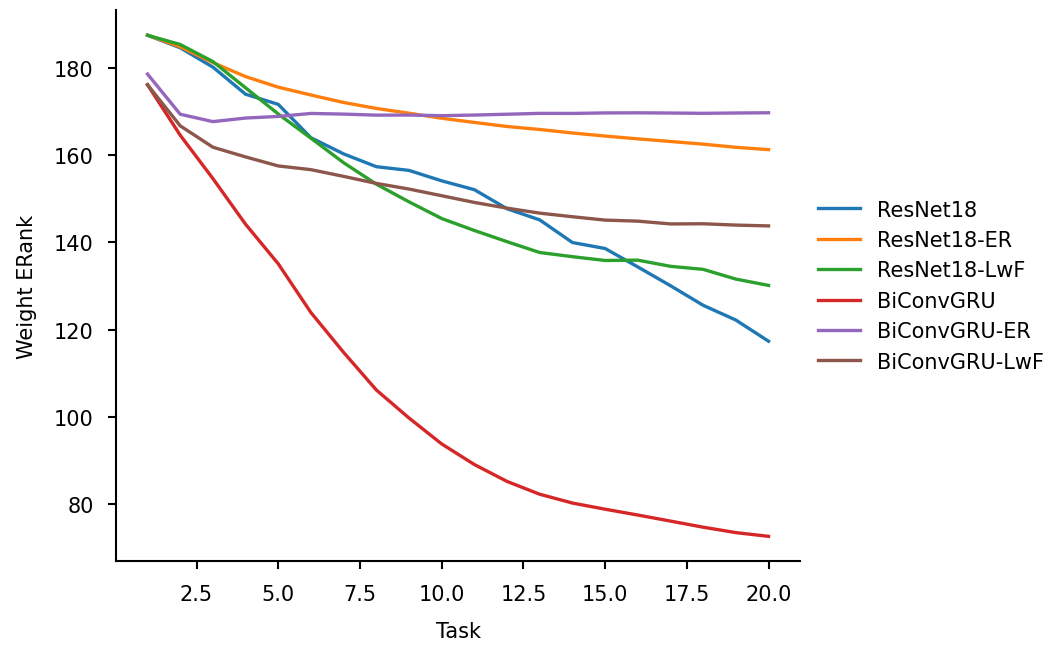}
\caption{The Weight eRank in Middle Layers For Split CIFAR-100}\label{fig:cifar-wer-2}
\end{subfigure}

\vspace{0.3cm}

\begin{subfigure}{0.48\linewidth}
\centering
\includegraphics[width=\linewidth]{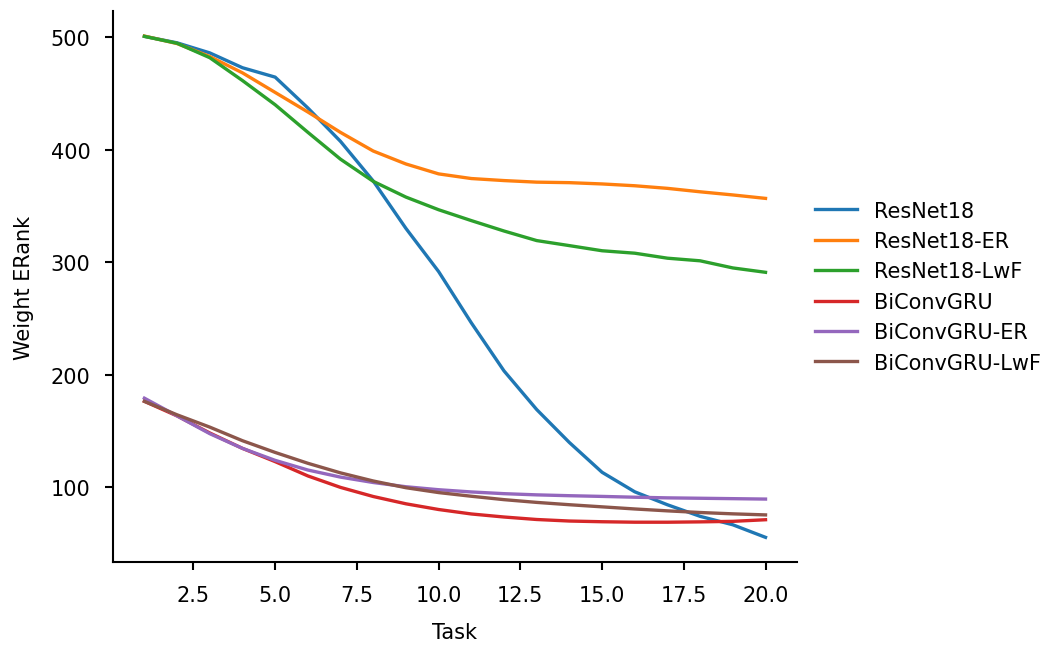}
\caption{The Weight eRank in Late Layers For Split CIFAR-100}\label{fig:cifar-wer-3}
\end{subfigure}
\hfill
\begin{subfigure}{0.48\linewidth}
\centering
\includegraphics[width=\linewidth]{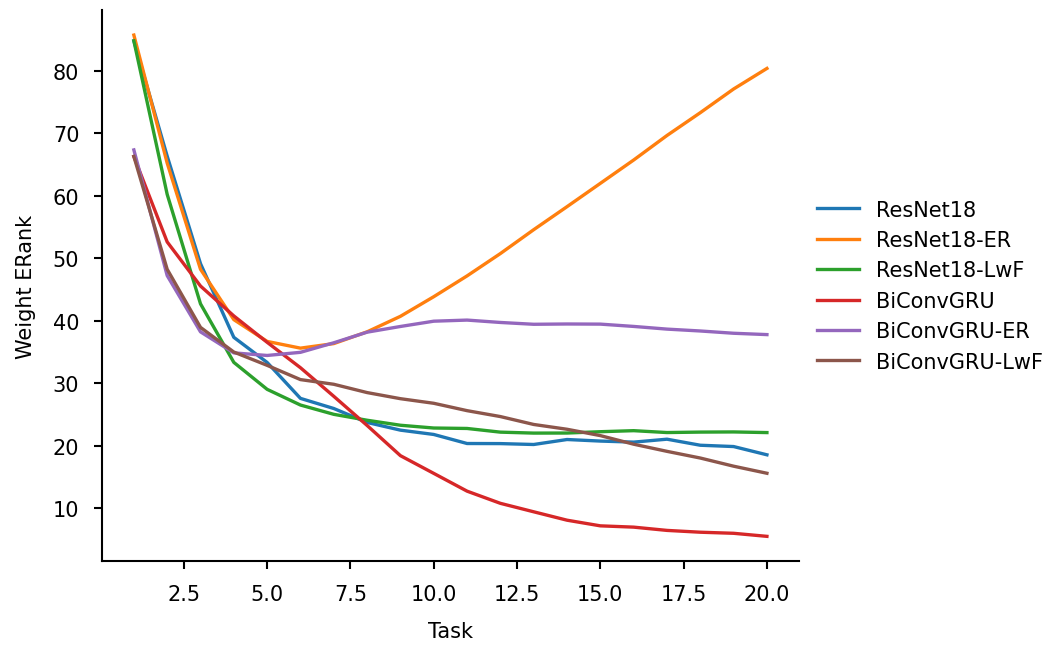}
\caption{The Weight eRank in Classification Layers For Split CIFAR-100}\label{fig:cifar-wer-head}
\end{subfigure}

\caption{For Split CIFAR-100, weight eRank indicates structural collapse in all layers for both ResNet-18 and Bi-ConvGRU. ER has the highest weight eRank and stabilizes internal structure, whereas LwF shows limited and architecture-dependent improvement.}
\label{fig:cifar-werank}

\end{figure*}

In general, pure architectures without any strategies lead to rapid decrease in accuracy, high forgetting, and sharp declines in activation and weight eRank. LwF is effective for stabilizing accuracy and moderating collapse, but its effect is limited and architecture-dependent. In contrast, ER shows the most significant improvement in maintaining higher accuracy, reducing forgetting, maintaining activation eRank, and slowing structural compression in weights. This effect applies to all four architectures. 

\section{Discussion}
\noindent This section combines our results from Split-MNIST and Split-CIFAR-100 with structural metrics and their peak-normalized values to explain the correlations between eRank collapse and forgetting across feed-forward and recurrent architectures. To be more specific, we explain why models forget, how different architectures shape collapse, and why CL strategies have different effectiveness. Overall, our findings show that catastrophic forgetting occurs when the model loses its degrees of freedom, as evidenced by the collapse in activations and weights. The effectiveness of CL strategies is strongly related to their capacity to deal with these collapses.

\begin{figure*}[t]
\centering

\begin{subfigure}{0.48\linewidth}
\centering
\includegraphics[width=\linewidth]{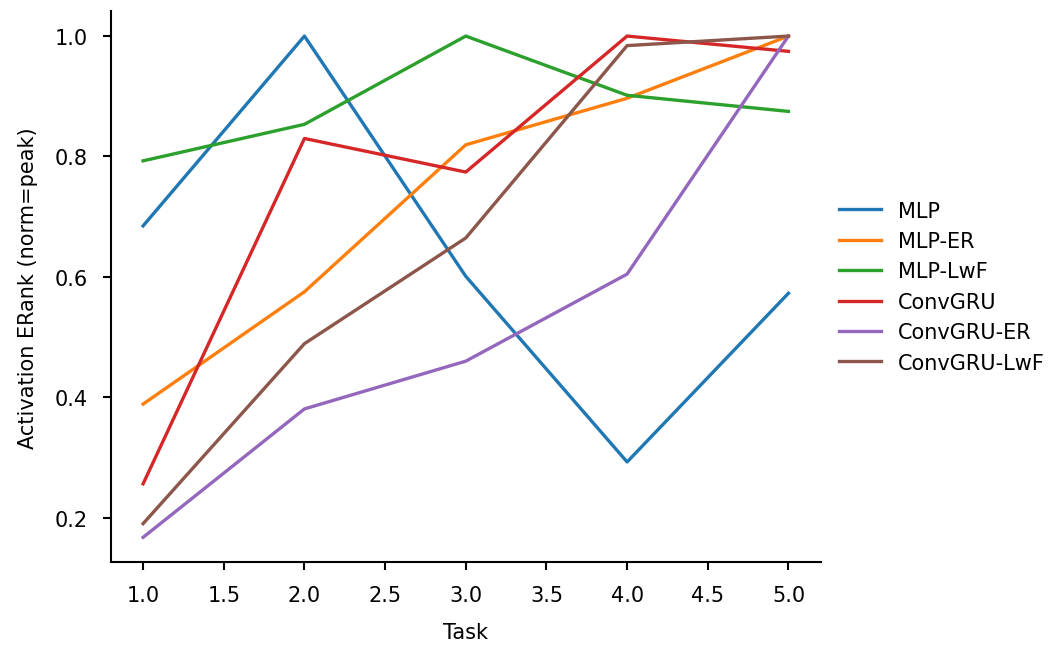}
  \caption{Peak-Normalized Activation eRank For Split MNIST \\MLP collapses rapidly due to unconstrained gradient propagation, leading to early weight and activation eRank collapse and severe forgetting.}
  \label{fig:mnist-normact}
\end{subfigure}
\hfill
\begin{subfigure}{0.48\linewidth}
\centering
\includegraphics[width=\linewidth]{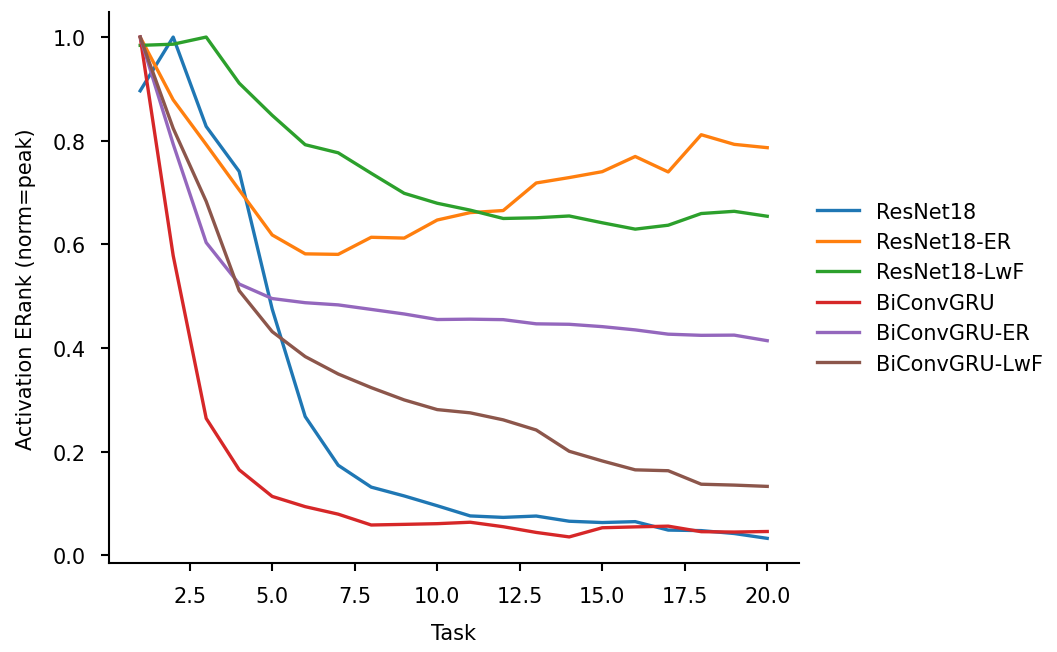}
\caption{Peak-Normalized Activation eRank For Split CIFAR-100}\label{fig:cifar-neract}
\end{subfigure}

\caption{Hidden layers map inputs to representations to identify the classes of tasks. When learning new tasks without effective regularization, old features are compressed into a low-dimensional subspace, which makes the decision boundaries non-separable. Therefore, activation eRank declines (i.e., suffers severe collapses) as forgetting increases. Under ER and LwF, eRanks are all stabilized; however, LwF shows limited improvement because the structural collapse happens, which further limits long-term learning and affects the activation eRank.}
\label{fig:mnist-werank}

\end{figure*}

\subsection{Representational Collapse Drives Forgetting}
\noindent The responsibility of hidden layers is to map raw input, pixels in our experiment, into representations that inputs from the same class cluster together while apart from those from different classes. When the network learns a new task without efficient regularization, the gradients push the shared weights to find new features, which forces the old features to shift and compress to a low-dimensional subspace. Therefore, the old decision boundaries are non-separable. Across all models in our study, activation eRank declines as performance degrades and forgetting increases. For MLP-SGD trained on Split-MNIST, the activation eRank increases as the first tasks are learned but collapses rapidly in Tasks 2–5. A similar trend appears in ResNet-18-SGD on Split CIFAR-100, as shown in peak-normalized curves in Figures~\ref{fig:mnist-normact} and~\ref{fig:cifar-neract}. ERank begins with a high value, which indicates that the model contains rich representations at early tasks, but collapses to almost zero in later tasks, along with the sharp increase in forgetting. In contrast, models with stabilized eRank, especially under the ER strategies, have very low forgetting.

Furthermore, the limited performance of LwF compared to ER suggests that stabilized eRank not only refers to no representational collapse. The MLP and ResNet-18 under LwF in Figures~\ref{fig:mnist-normact} and~\ref{fig:cifar-neract} show that eRank remains stable, since the purpose of this strategy is to keep the network’s output function stable and the features recognizable. However, it does not place any direct penalty on the weight matrices $\mathbf{W}$, so the weight matrices still experience structural collapse compared to the ER. This makes the network less capable of long-term learning compared to ER, which reconstructs old feature directions through replay. Figures~\ref{fig:mnist-ner-1} and~\ref{fig:cifar-ner-1} are such evidence, which indicates that the features are extracted in fewer directions and the network does not have a rich and diverse set of representations from the beginning. More severe structural collapse in the middle layers and the late layers (Figures~\ref{fig:mnist-ner-2},~\ref{fig:cifar-ner-2} and~\ref{fig:cifar-ner-3}) also reduces the network’s ability to perform complex and non-linear combinations between features from the early layers and the current learning task, which further generates less rich variety of high-level representations. LwF also performs worse in classification layers (Figures~\ref{fig:mnist-ner-head} and~\ref{fig:cifar-ner-head}), showing that the network is less effective in preventing decision boundary drift. Therefore, structural collapse reduces plasticity, thereby causing activation collapse and leading to catastrophic forgetting.

\begin{figure*}[!t]
\centering

\begin{subfigure}{0.48\linewidth}
\centering
\includegraphics[width=\linewidth]{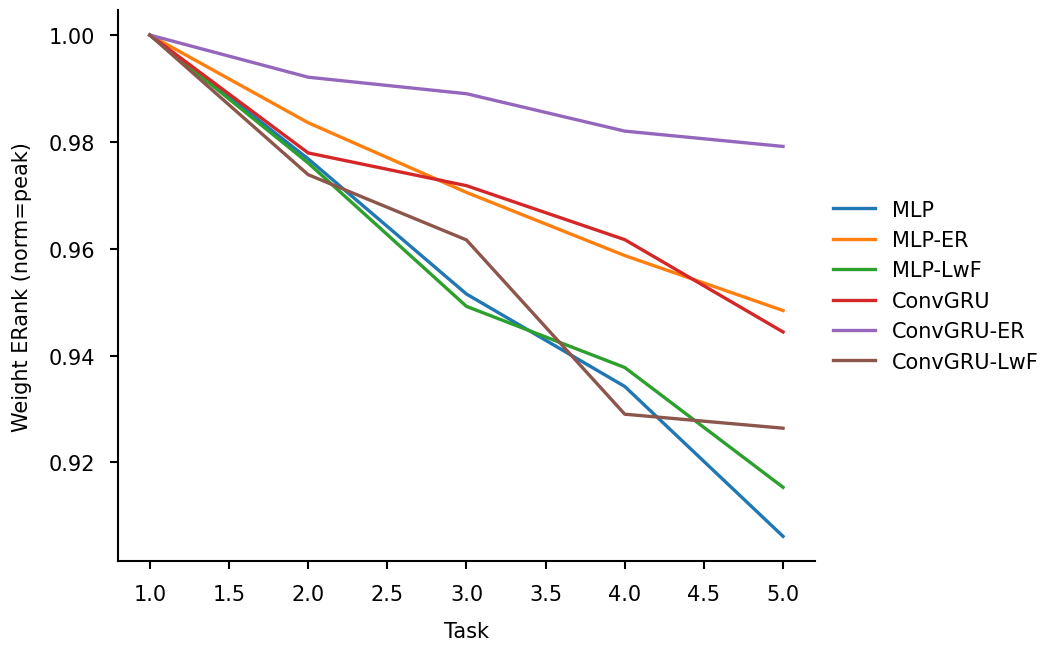}
\caption{Peak-Normalized Weight eRank in Early Layers For Split MNIST}\label{fig:mnist-ner-1}
\end{subfigure}
\hfill
\begin{subfigure}{0.48\linewidth}
\centering
\includegraphics[width=\linewidth]{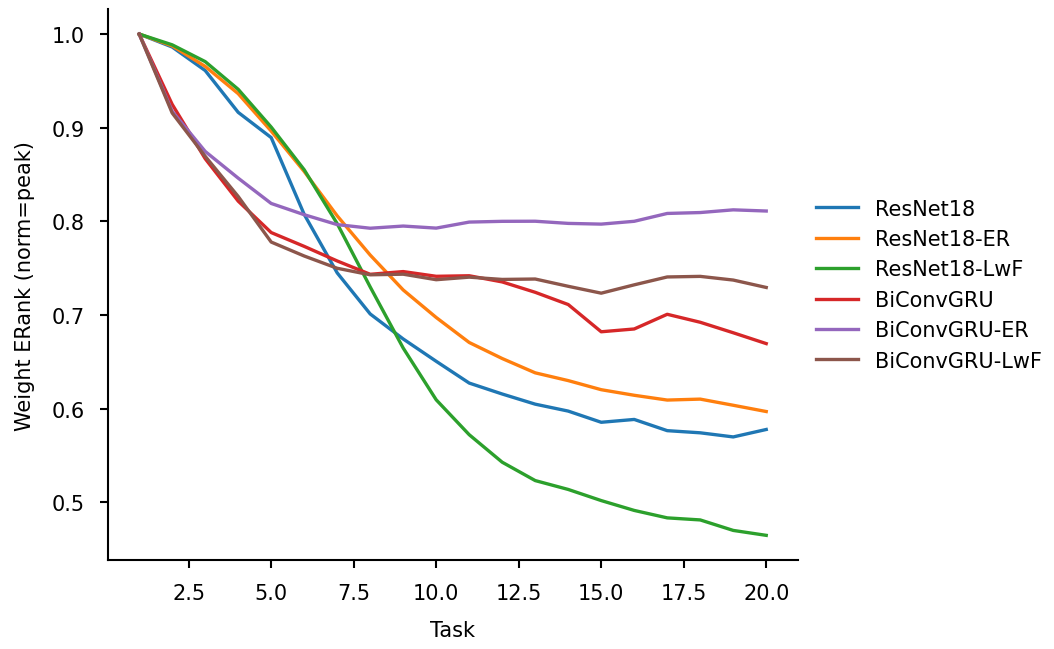}
\caption{Peak-Normalized Weight eRank in Early Layers For Split CIFAR-100}\label{fig:cifar-ner-1}
\end{subfigure}

\vspace{0.3cm}

\begin{subfigure}{0.48\linewidth}
\centering
\includegraphics[width=\linewidth]{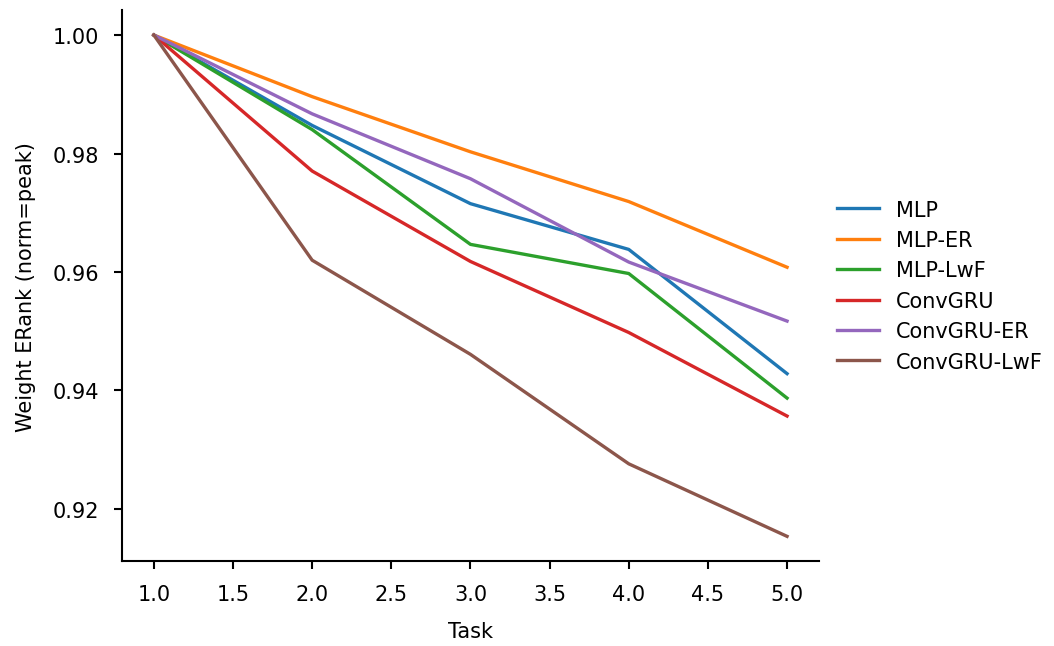}
\caption{Peak-Normalized Weight eRank in Middle Layers For Split MNIST}\label{fig:mnist-ner-2}
\end{subfigure}
\hfill
\begin{subfigure}{0.48\linewidth}
\centering
\includegraphics[width=\linewidth]{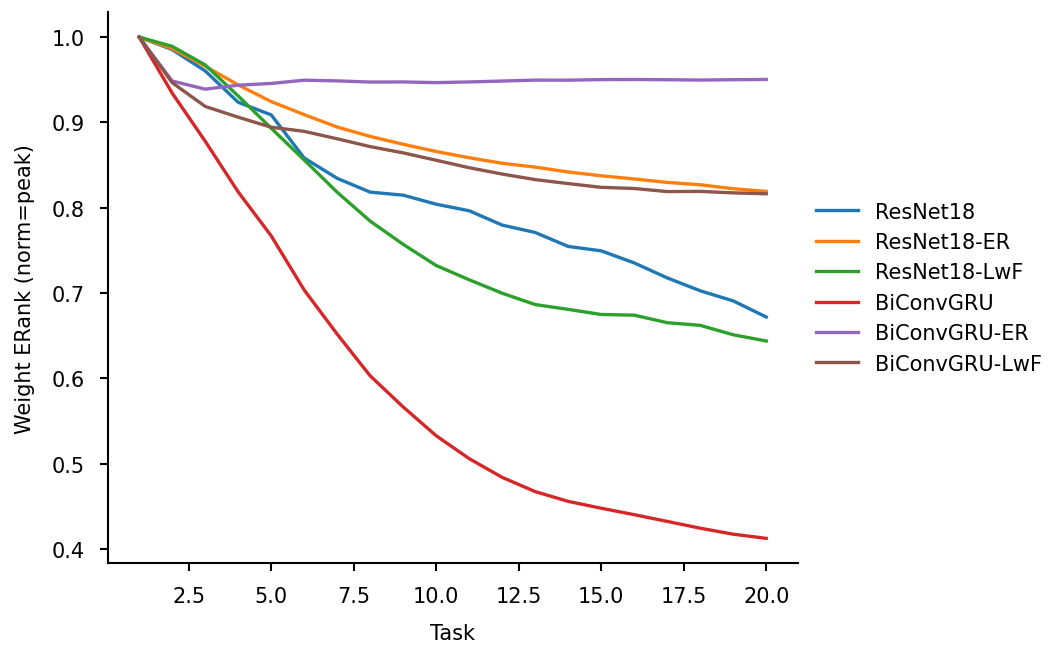}
\caption{Peak-Normalized Weight eRank in Middle Layers For Split CIFAR-100}\label{fig:cifar-ner-2}
\end{subfigure}

\vspace{0.3cm}

\begin{subfigure}{0.48\linewidth}
\centering
\includegraphics[width=\linewidth]{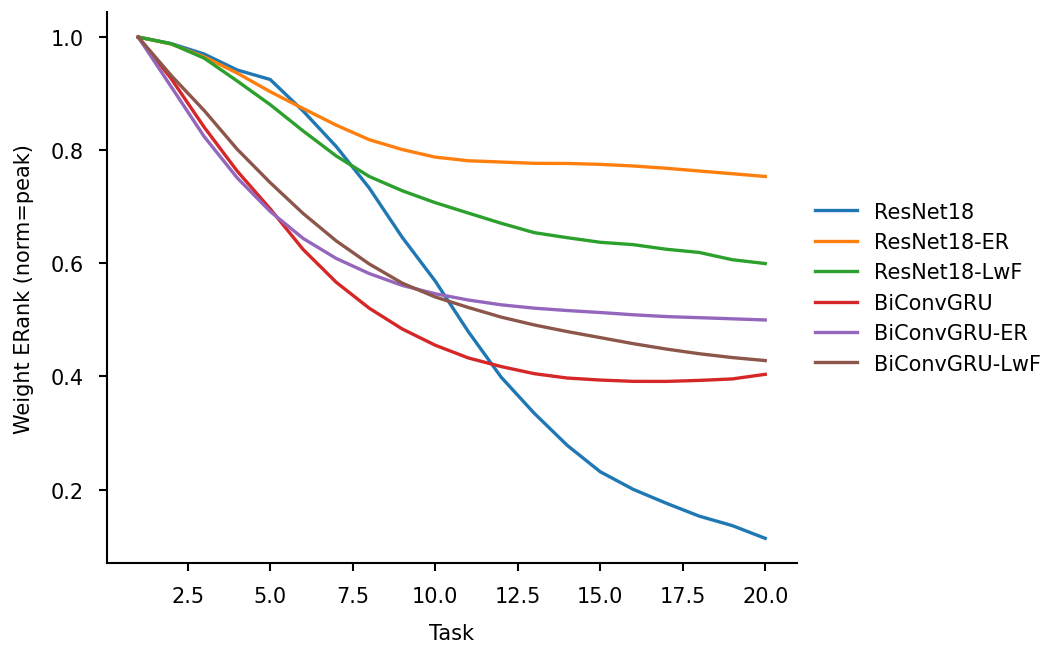}
\caption{Peak-Normalized Weight eRank in Late Layers For Split CIFAR-100\\ConvGRU and Bi-ConvGRU use recurrent gating to reduce gradient interference and delay structural failure, but this comes at the cost of aggressive early compression that limits representational richness and long-term capacity. ResNet-18 initially benefits from skip connections, which stabilize training and delay collapse (as shown in the plots above). Still, as tasks accumulate, it collapses, indicating that residual connections can no longer preserve plasticity. }\label{fig:cifar-ner-3}
\end{subfigure}
\hfill
\begin{subfigure}{0.48\linewidth}
\centering
\includegraphics[width=\linewidth]{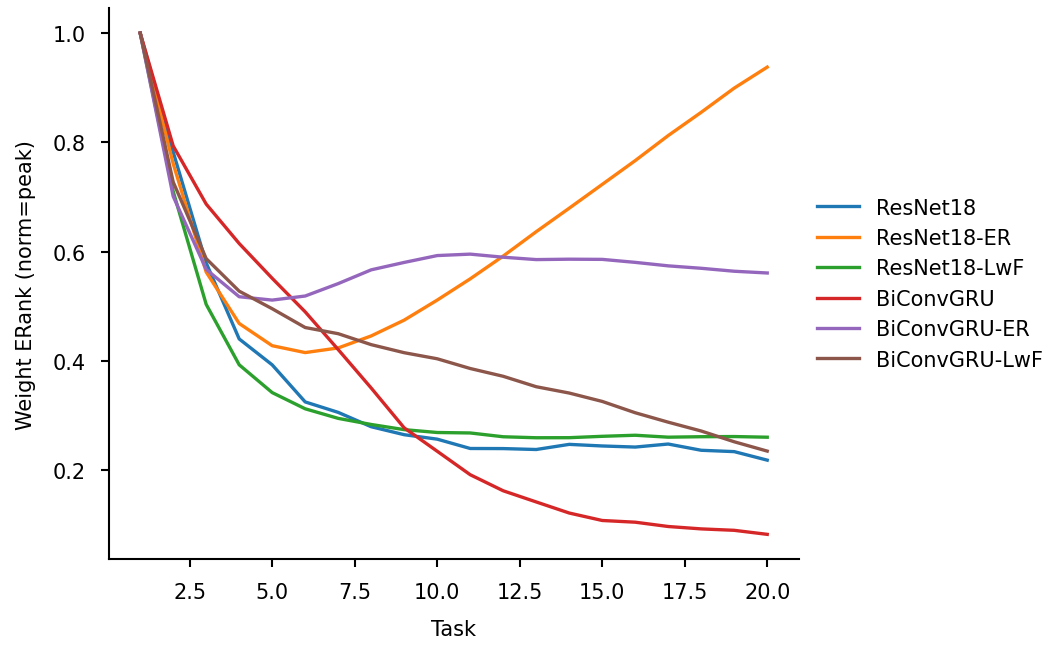}
\caption{Peak-Normalized Weight eRank in Classification Layers For Split CIFAR-100}\label{fig:cifar-ner-head}
\end{subfigure}
\end{figure*}

\begin{figure*}[!t]\ContinuedFloat
\centering
\begin{subfigure}{0.48\linewidth}
\centering
\includegraphics[width=\linewidth]{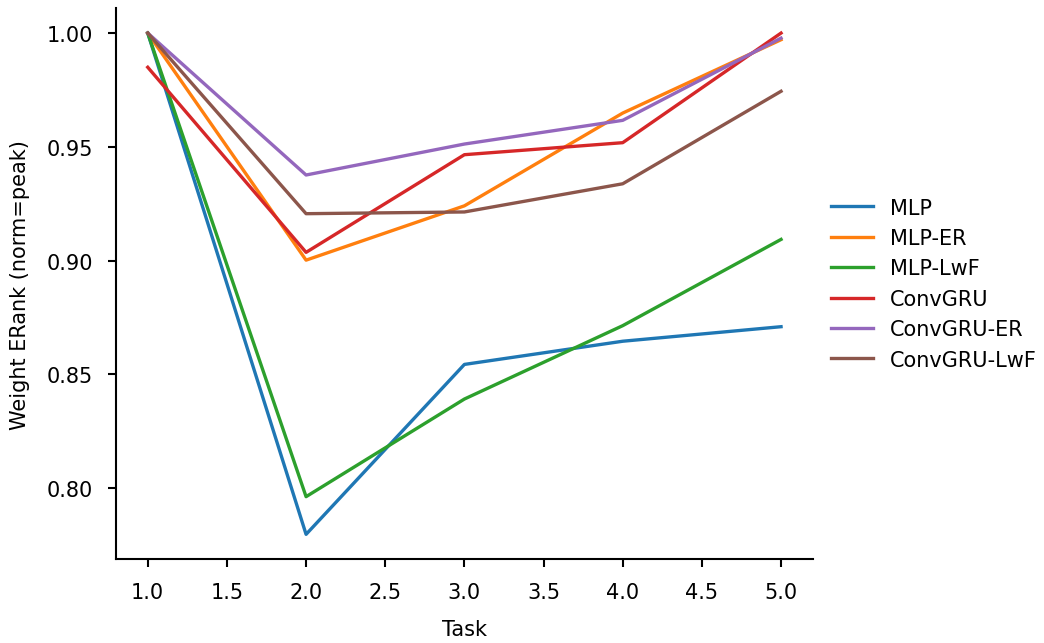}
\caption{Peak-Normalized Weight eRank in Classification Layers For Split MNIST}\label{fig:mnist-ner-head}
\end{subfigure}

\caption{The plots show that structural collapse drives forgetting. Starting in the early layers, features are extracted along fewer directions under SGD and LwF than under ER (especially SGD), thereby having more limited representational diversity. Such structural collapse becomes more severe in the middle and late layers since the network loses the ability to form rich, high-level features. This degradation eventually reaches the classification layers, causing severe decision boundary drift. Moreover, the plots indicate that different architectures fail in different ways.}
\label{fig:mnist-cifar-ner}

\end{figure*}

\subsection{Different Architectures Shape Collapse}
\noindent Although collapse is the same, different architectures fail in distinct ways because their architectural biases determine how gradients flow and where structural collapse is initiated. The MLP, used in the Split-MNIST environment, is the baseline for investigating the direct correlation between collapse and catastrophic forgetting. This architecture is very simple that does not include spatial convolutions, recurrence gates, and normalization, which offers minimal protection against overwriting of gradients. The fully connected layers in MLP indicate that gradient updates propagate across all weights simultaneously. Because it does not have structural gates or normalization, its weight subspace collapses rapidly, resulting in a sharp decline in weight eRank. The model also explores a high-dimensional activation space at the beginning of the training, as seen in Figure~\ref{fig:mnist-normact}, but the activation eRank drops immediately. This is consistent with the above analysis that the loss of feature separability is the immediate cause of forgetting.

The ConvGRU, which contains convolutional layers and a GRU, fundamentally makes an effort to reduce the collapse observed in the MLP. The purpose of recurrence is to maintain a compressed history of past knowledge. Different with the MLP, ConvGRU’s activation eRank is lower since the recurrent gates effectively compress the representations into a minimal feature subspace. Therefore, peak-normalized eRank is needed to understand its structure and compare it with other architectures. Due to its recurrent gates, the model can reduce gradient interference that would otherwise destroy old features. This helps ensure that the eRank weight does not collapse immediately and avoids catastrophic structural failure in MLP. However, this aggressive early compression restricts usable representations, reducing the models' ability to achieve higher learning capacity. The model also lacks sufficient structural capacity to learn all the required fine-grained features. These are the reasons for the lower peak-normalized values in Figure~\ref{fig:mnist-ner-2} than in MLP. The Bi-ConvGRU extends to deal with the Split CIFAR-100, so the mechanism behind is similar to ConvGRU. In Figure~\ref{fig:cifar-ner-2}, the weight collapse in the middle layers is accelerated since the gradients repeatedly strengthen the shared filters, which forces it to converge. Then, for the peak-normalized weight eRank in the recurrent stage (Figure~\ref{fig:cifar-ner-3}), the curve is flatter than that of ResNet-18. This is not due to preserved stability, but because the constrained feature map lacks the dimensional richness required for classification, which is further proved by Figure~\ref{fig:cifar-ner-head}. This forces the network to give up some of its expressive power.

The ResNet-18 architecture trained on the more challenging Split-CIFAR-100 dataset under the Class-IL setting is the baseline of purely feedforward models with high capacity. Although residual skip connections enable the training of very deep networks, the benefit is temporary. Skip connections successfully enrich early representations and reduce the magnitude of gradient updates, which delays early collapse as seen in the MLP. However, the collapse occurs suddenly and permanently as tasks accumulate. The activation eRank approaches zero in Figure~\ref{fig:cifar-neract}, demonstrating a severe representational collapse where all the distinguished features are destroyed. In addition, the weight of eRank drops sharply in the late layers, representing the loss of plasticity. These collapses show that while skip connections are effective at preventing catastrophic updates at the beginning, they do not preserve the required features during long-term training. 

\hfill \break
\subsection{Effectiveness of Continual Learning Strategies} 
\noindent Across all experimental models, ER is the most effective strategy for reducing both forgetting and collapse in our study. ER continuously replays information from previous tasks during learning on new tasks. In Split-MNIST, the ER models preserve or increase activation eRank over time, which demonstrates that replay maintains a rich feature subspace that contains both old and new knowledge. Similarly, for the weight eRank curves, ER significantly slows the collapse in the early and middle layers. The peak-normalized eRank curves of the classification layers indicate that the replay helps preserve the decision boundaries. Results on Split CIFAR-100 are in the same pattern. Both architectures under ER enable continual representational growth even in the late layers. ER shows its ability to delay the sharp collapse in SGD and LwF, thereby flattening the decline and increasing the peak-normalized ranks. In general, the plots indicate that ER is uniquely effective. For activation eRank, the network is forced to find the representations that enable the features’ separability for both old and new tasks, which slows representational collapse by maintaining the quality of representations. For the weight eRank, the mixed gradient is generated through incorporating the signals from the old tasks to constrain the effects of the destabilizing gradient, which ensures that the network keeps a high number of active degrees of freedom in its parameter space to preserve its flexibility for future tasks.

As discussed above, LwF has a consistent but limited ability to reduce forgetting. Its improvements and limitations are both shown in the results. In Split-MNIST, accuracy plots show that LwF reduces forgetting in the first four tasks, but a late clear collapse in Task 5, while ConvGRU-LwF has a better accuracy but not as strong as ER. These behaviors are consistent with the activation eRank curves, which indicate that this strategy prefers preservation rather than enrichment. The weight eRank curves clearly show such limitations, especially when dealing with feed-forward networks. In the early and middle layers of MLPs, LwF does not prevent structural collapse, even compared to SGD in some cases. The Peak-normalized eRank curves in the classifier further show that LwF improves stability, but the classification subspaces remain narrow. Thus, while LwF protects output behavior, it fails to solve the problem of internal capacity erosion. The results of the Split CIFAR-100 dataset further evidence this. The activation eRank curves show that while LwF prevents catastrophic collapse, it does not restore representational richness. For ResNet-18, LwF experiences the most severe weight collapse as MLP did before. Overall, the plots show that LwF reduces forgetting through functional regularization but does not preserve plasticity. It does not protect the corresponding feature space while preserving output behavior, so models retain performance shortly before gradually losing the representational depth needed to handle old tasks.

\subsection{Limitations and Future Work}
\noindent Although our study explains how catastrophic forgetting relates to collapse from different perspectives, limitations exist and restrict further findings. The datasets are simple compared to the complexity of continual learning in the real world. Split-MNIST and Split CIFAR-100 clearly show sequential learning, but both are fixed and stationary, which lack data drift, domain shift, and overlapped categories~\cite{ddddiscc}. Thus, while our study observes the dynamics of collapse, it is unclear whether the observed relationships can also be applied to complex domains such as natural video streams or non-stationary reinforcement learning settings.

Further, the diversity of our architectures is limited. The comparison includes MLP, ConvGRU, ResNet-18, and Bi-ConvGRU, but lacks evaluation of major groups, such as attention-based models and modular networks. These architectures work differently in how they preserve representations. Our findings might not apply to such models. Future work may investigate collapse dynamics in transformer-based architectures, modular networks, and memory-enhanced models to test whether the conclusions apply to other architectures.

In addition, the GRU designs are simple, which limits our findings of correlations in recurrence. We also focus only on supervised learning, where the labels are known to the models. Previous work suggests that self-supervision creates rich features~\cite{fini2022selfsupervisedmodelscontinuallearners}~\cite{ericsson2021selfsupervisedmodelstransfer}, which might show different collapse behaviors. Therefore, it may be meaningful to explore how self-supervised and unsupervised objectives affect representational and structural collapse in continual learning.

Finally, our study mainly relies on effective rank, which measures effective dimensionality. However, we do not use other metrics to further verify our results, such as subspace alignment that check whether the model is pointing correctly, or functional capacity.  Future work may incorporate other metrics, especially gradient and Fisher-based measures mentioned above, to further validate and extend the insights from effective rank and catastrophic forgetting.
\section{Conclusion}
\noindent Our study investigates the correlation between catastrophic forgetting and collapse by measuring effective rank. Since networks gradually lose the dimensionalities required to learn new knowledge, we view forgetting as a geometric failure. To study this, we evaluated four architectures, including MLP, ConvGRU, ResNet-18, and Bi-ConvGRU, on Split MNIST and Split CIFAR-100 datasets. Those models are trained using SGD, Learning without Forgetting, and Experience Replay. We also used effective rank to measure the representational richness for both weights and activations. The results consistently show that forgetting and collapse are strongly related, and continual learning strategies help models preserve both capacity and performance in different efficiency.

Our experiments provided clear evidence of the correlation across all settings. The declines in both activation and weight eRank are closely related to the performance degradation. When the eRank collapses, the model runs out of space to store new knowledge, and accuracy decreases accordingly. This strong alignment between eRank trends and forgetting curves suggests that representational and structural collapse are the main drivers of catastrophic forgetting. Furthermore, our experiments tested two groups of architectures: feed-forward and recurrent models, which show that architectures shape the trajectory of collapse. Feedforward models, such as MLP and ResNet-18, start with high representational capacity, but collapse rapidly without CL strategies. The residual connections in ResNet-18 delay this process through stabilizing early representations, but this benefit is temporary. When training begins on new tasks, both activation and weight eRank drop sharply, further leading to a permanent loss of plasticity. Recurrent models behave differently because of their gating mechanisms. They compress the representations at an earlier stage, thereby stabilizing training and delaying forgetting. However, the trajectory of eRank indicates that this early compression also limits representational richness and further reduces the long-term capacity. Thus, recurrent mechanism trades capacity for stability.

We also compared three learning strategies and found that their success depends primarily on their ability to handle collapse, further supporting the correlation. The vanilla SGD, which serves as a baseline, fails since it causes the immediate and severe collapse in both the representation and the structure. LwF improves the stability through constraining outputs, but our study shows that this fails to preserve internal capacity. Although activation eRank under LwF remains stable during training, weight eRank continues to decline in all layers. This occurs because functional behavior is preserved, whereas the feature space degrades, which also explains why LwF can delay forgetting but cannot keep the ability for long-term learning.

In contrast, ER is the most effective strategy that preserves both performance and functionality. For both architectures, ER effectively slows down the collapses. The eRank curves indicate that it maintains richer feature subspaces and stabilizes decision boundaries. These results demonstrate that solving catastrophic forgetting requires not only stabilizing outputs, but also preserving representational and structural capacity.

\bibliographystyle{IEEEtran}
\bibliography{ref}

\begin{thebibliography}{10}
\providecommand{\url}[1]{#1}
\csname url@samestyle\endcsname
\providecommand{\newblock}{\relax}
\providecommand{\bibinfo}[2]{#2}
\providecommand{\BIBentrySTDinterwordspacing}{\spaceskip=0pt\relax}
\providecommand{\BIBentryALTinterwordstretchfactor}{4}
\providecommand{\BIBentryALTinterwordspacing}{\spaceskip=\fontdimen2\font plus
\BIBentryALTinterwordstretchfactor\fontdimen3\font minus \fontdimen4\font\relax}
\providecommand{\BIBforeignlanguage}[2]{{%
\expandafter\ifx\csname l@#1\endcsname\relax
\typeout{** WARNING: IEEEtran.bst: No hyphenation pattern has been}%
\typeout{** loaded for the language `#1'. Using the pattern for}%
\typeout{** the default language instead.}%
\else
\language=\csname l@#1\endcsname
\fi
#2}}
\providecommand{\BIBdecl}{\relax}
\BIBdecl

\bibitem{wang2024comprehensivesurveycontinuallearning}
L.~Wang, X.~Zhang, H.~Su, and J.~Zhu, ``A comprehensive survey of continual learning: Theory, method and application,'' 2024.

\bibitem{Kirkpatrick_2017}
J.~Kirkpatrick, R.~Pascanu, N.~Rabinowitz, J.~Veness, G.~Desjardins, A.~A. Rusu, K.~Milan, J.~Quan, T.~Ramalho, A.~Grabska-Barwinska, D.~Hassabis, C.~Clopath, D.~Kumaran, and R.~Hadsell, ``Overcoming catastrophic forgetting in neural networks,'' \emph{Proceedings of the National Academy of Sciences}, vol. 114, no.~13, p. 3521–3526, Mar. 2017.

\bibitem{mallya2018packnetaddingmultipletasks}
A.~Mallya and S.~Lazebnik, ``Packnet: Adding multiple tasks to a single network by iterative pruning,'' 2018.

\bibitem{rolnick2019experiencereplaycontinuallearning}
D.~Rolnick, A.~Ahuja, J.~Schwarz, T.~P. Lillicrap, and G.~Wayne, ``Experience replay for continual learning,'' 2019.

\bibitem{nature2024}
S.~Dohare, J.~Hernandez-Garcia, Q.~Lan, P.~Rahman, A.~Mahmood, and R.~Sutton, ``Loss of plasticity in deep continual learning,'' \emph{Nature}, vol. 632, pp. 768--774, 08 2024.

\bibitem{effrank}
O.~Roy and M.~Vetterli, ``The effective rank: A measure of effective dimensionality,'' 01 2007.

\bibitem{setting}
H.~Hihn and D.~Braun, ``Hierarchically structured task-agnostic continual learning,'' \emph{Machine Learning}, vol. 112, 12 2022.

\bibitem{10.3389/fnbot.2021.658280}
H.~Duan, P.~Wang, Y.~Huang, G.~Xu, W.~Wei, and X.~Shen, ``Robotics dexterous grasping: The methods based on point cloud and deep learning,'' \emph{Frontiers in Neurorobotics}, vol. Volume 15 - 2021, 2021.

\bibitem{ilthreetypes}
G.~van~de Ven, T.~Tuytelaars, and A.~Tolias, ``Three types of incremental learning,'' \emph{Nature Machine Intelligence}, vol.~4, pp. 1--13, 12 2022.

\bibitem{liu2023teamworkgoodempiricalstudy}
M.~Liu and L.~Huang, ``Teamwork is not always good: An empirical study of classifier drift in class-incremental information extraction,'' 2023.

\bibitem{li2017learningforgetting}
Z.~Li and D.~Hoiem, ``Learning without forgetting,'' 2017.

\bibitem{lopezpaz2022gradientepisodicmemorycontinual}
D.~Lopez-Paz and M.~Ranzato, ``Gradient episodic memory for continual learning,'' 2022.

\bibitem{chaudhry2019efficientlifelonglearningagem}
A.~Chaudhry, M.~Ranzato, M.~Rohrbach, and M.~Elhoseiny, ``Efficient lifelong learning with a-gem,'' 2019.

\bibitem{Chaudhry_2018}
A.~Chaudhry, P.~K. Dokania, T.~Ajanthan, and P.~H.~S. Torr, \emph{Riemannian Walk for Incremental Learning: Understanding Forgetting and Intransigence}.\hskip 1em plus 0.5em minus 0.4em\relax Springer International Publishing, 2018, p. 556–572.

\bibitem{zenke2017continuallearningsynapticintelligence}
F.~Zenke, B.~Poole, and S.~Ganguli, ``Continual learning through synaptic intelligence,'' 2017.

\bibitem{saha2021spacestructuredcompressionsharing}
G.~Saha, I.~Garg, A.~Ankit, and K.~Roy, ``Space: Structured compression and sharing of representational space for continual learning,'' 2021.

\bibitem{Rumelhart1986LearningRB}
D.~E. Rumelhart, G.~E. Hinton, and R.~J. Williams, ``Learning representations by back-propagating errors,'' \emph{Nature}, vol. 323, pp. 533--536, 1986.

\bibitem{he2015deepresiduallearningimage}
K.~He, X.~Zhang, S.~Ren, and J.~Sun, ``Deep residual learning for image recognition,'' 2015.

\bibitem{conv222}
N.~Ballas, L.~Yao, C.~Pas, and A.~Courville, ``Delving deeper into convolutional networks for learning video representations,'' 11 2015.

\bibitem{10.5555/2969239.2969329}
X.~Shi, Z.~Chen, H.~Wang, D.-Y. Yeung, W.-k. Wong, and W.-c. Woo, ``Convolutional lstm network: a machine learning approach for precipitation nowcasting,'' ser. NIPS'15.\hskip 1em plus 0.5em minus 0.4em\relax Cambridge, MA, USA: MIT Press, 2015, p. 802–810.

\bibitem{li2016videolstmconvolvesattendsflows}
Z.~Li, E.~Gavves, M.~Jain, and C.~G.~M. Snoek, ``Videolstm convolves, attends and flows for action recognition,'' 2016.

\bibitem{pfülb2019comprehensiveapplicationorientedstudycatastrophic}
B.~Pfülb and A.~Gepperth, ``A comprehensive, application-oriented study of catastrophic forgetting in dnns,'' 2019.

\bibitem{ddddiscc}
G.~van~de Ven, T.~Tuytelaars, and A.~Tolias, ``Three types of incremental learning,'' \emph{Nature Machine Intelligence}, vol.~4, pp. 1--13, 12 2022.

\bibitem{fini2022selfsupervisedmodelscontinuallearners}
E.~Fini, V.~G.~T. da~Costa, X.~Alameda-Pineda, E.~Ricci, K.~Alahari, and J.~Mairal, ``Self-supervised models are continual learners,'' 2022.

\bibitem{ericsson2021selfsupervisedmodelstransfer}
L.~Ericsson, H.~Gouk, and T.~M. Hospedales, ``How well do self-supervised models transfer?'' 2021.

\end{thebibliography}

\end{document}